\documentclass[10pt,twocolumn,letterpaper]{article}

\usepackage[accsupp]{axessibility}  

\usepackage[pagenumbers]{cvpr} 

\usepackage{bm}
\usepackage[export]{adjustbox}

\usepackage{graphicx}
\usepackage{subcaption}
\usepackage{float} 
\usepackage{array}
\usepackage{longfigure}

\definecolor{cvprblue}{rgb}{0.21,0.49,0.74}
\usepackage[pagebackref,breaklinks,colorlinks,allcolors=cvprblue]{hyperref}
\usepackage{bibunits}
\def\paperID{14208} 
\def\confName{CVPR}
\def\confYear{2026}

\title{ATATA: One Algorithm to Align Them All}

\author{
    Boyi Pang\textsuperscript{1,*}
    \quad{}Savva Ignatyev\textsuperscript{2,*}
    \quad{}Vladimir Ippolitov\textsuperscript{2,*}
    \quad{}Ramil Khafizov\textsuperscript{2}
    \\\
    Yurii Melnik\textsuperscript{2}
    \quad{}Oleg Voynov\textsuperscript{2,3}
    \quad{}Maksim Nakhodnov\textsuperscript{4,5,6}
    \quad{}Aibek Alanov\textsuperscript{4,7}
    \\\
    Xiaopeng Fan\textsuperscript{1,8,9,$\dagger$}
    \quad{}Peter Wonka\textsuperscript{10,$\dagger$}
    \quad{}Evgeny Burnaev\textsuperscript{1,2,3}
    \medskip\\\
    \textsuperscript{1}Harbin Institute of Technology
    \quad{}\textsuperscript{2}Applied AI Institute
    \quad{}\textsuperscript{3}AXXX
    \\\
    \textsuperscript{4}FusionBrain Lab, AXXX
    \quad{}\textsuperscript{5}MSU
    \quad{}\textsuperscript{6}Constructor University
    \quad{}\textsuperscript{7}HSE University
    \\\
    \textsuperscript{8}Peng Cheng Laboratory
    \quad{}\textsuperscript{9}HIT Suzhou Research Institute
    \quad{}\textsuperscript{10}KAUST
    \medskip\\\
    \textsuperscript{*}Equal contribution
    \quad{}\textsuperscript{$\dagger$}Indicates the corresponding author
}

\begin{document}
\twocolumn[{%
  \renewcommand\twocolumn[1][]{#1}%
  \maketitle
  \begin{center}
    \vspace{-1.2em} 
    \includegraphics[width=\textwidth]{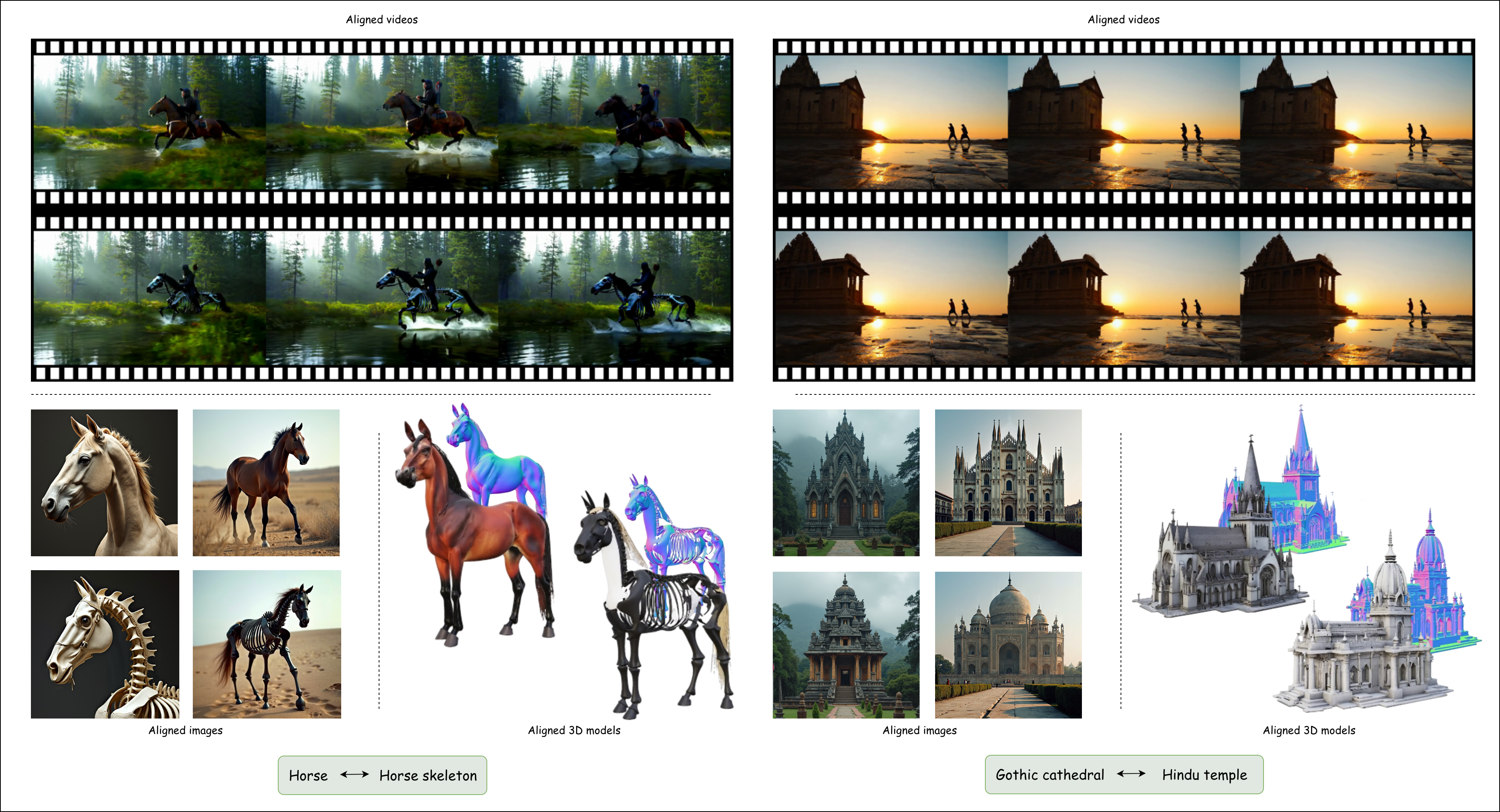}
    \captionof{figure}{Visualization of generated images, videos, and 3D shapes using our method. The left pair is (horse animal, horse skeleton), the right pair is (gothic temple, Hindu temple).}
    \label{fig:teaser}
    \vspace{0.9em}
  \end{center}
}]
\begin{abstract}
 We suggest a new multi-modal algorithm for joint inference of paired structurally aligned samples with Rectified Flow models.
 While some existing methods propose a codependent generation process, they do not view the problem of joint generation from a structural alignment perspective.
 Recent work uses Score Distillation Sampling to generate aligned 3D models, but SDS is known to be time-consuming, prone to mode collapse, and often provides cartoonish results. 
 By contrast, our suggested approach relies on the joint transport of a segment in the sample space, yielding faster computation at inference time.
 Our approach can be built on top of an arbitrary Rectified Flow model operating on the structured latent space. 
 We show the applicability of our method to the domains of image, video, and 3D shape generation using state-of-the-art baselines and evaluate it against both editing-based and joint inference-based competing approaches. 
 We demonstrate a high degree of structural alignment for the sample pairs obtained with our method and a high visual quality of the samples.
 Our method improves the state-of-the-art for image and video generation pipelines. 
 For 3D generation, it is able to show comparable quality while working orders of magnitude faster.
\href{https://voyleg.github.io/atata/}{voyleg.github.io/atata/}
\end{abstract}    
\section{Introduction}
\label{sec:intro}


Images, videos, and 3D models are three important domains for generative AI.
While users often employ AI to generate one sample at a time, there are many use cases where a user wants to generate a set of examples that are somehow related.
Examples are personalization~\cite{avrahami2023chosen}, where a user wants to have the same object or person appearing in multiple images or videos, or style consistent generation~\cite{Hertz_2024_CVPR}, where multiple samples should be generated in the same style.

In this paper, we address the topic of structurally aligned generation.
The goal of structurally aligned generation is to generate multiple samples (images, videos, 3D models) that showcase different main objects, scenes, or environments, but with their semantically (structurally) corresponding parts aligned across spatial/temporal dimensions.
This aligned generation comes in handy in generating virtual worlds for training and entertainment, where different objects and scene parts can be easily replaced across domains and "sewn up" into a new scene. 
It can also help in CAD and CAM applications for creating objects with interchangeable parts.
For image and video editing, these tools can be useful for generating artistic effects, for example, transitions called matchcuts~\cite{Pardo2025MatchDiffusion}. 
Finally, aligned generation methods can be useful for paired synthetic training data generation for image, video, and 3D, which can be later used to train editing models. 
While existing methods try to use editing for aligned generation~\cite{qwen_image_edit_2025, mvedit2024, vace}, this biases the generation to construct samples that fit one of the descriptions better than the others.

Current methods for aligned generation fall into three categories: 1) Editing-based methods, which use the input sample defined by the first description and force the second sample to adopt its structure.
2) Zero-shot generation using large foundation models. Many large generative models like Nanobanana~\cite{nanobanana}, QWEN~\cite{wu2025qwenimagetechnicalreport}, or FLUX~\cite{blackforestlabs_flux1dev_2024} can generate a grid of semi-consistent images or 3D renders.  
3) Native aligned generation. Some works solve the problem of the joint aligned generation, introducing some kind of interaction between the samples during generation. 
This includes joint generation up to some step in MatchDiffusion~\cite{Pardo2025MatchDiffusion}, attention sharing~\cite{Hertz_2024_CVPR}, or embedding into a common latent space in A3D~\cite{Ignatyev2025A3D}.

We propose a novel, highly generalizable, multi-modal method for structurally aligned generation. 
Inspired by A3D~\cite{Ignatyev2025A3D}, we rethink the necessary properties of transitions between samples with regard to rectified flow field numerical integration and introduce additional constraints to the process, which are necessary for the algorithm's convergence.
The method is theoretically applicable to any rectified flow model that operates on the structured latent space, which is demonstrated for images, videos, and 3D objects with minor implementation differences between them. 
The method requires only changing the inference loop of the flow model, without changing any other components of the pipeline, and does not depend on the specific domain. 
The main idea of the method is to perform joint inference for pairs of samples by moving them and the linear interpolations between them along the velocity field of the rectified flow model while preserving the interpolation structure. 
We additionally introduce joint co-guidance during the transport process aimed at ensuring the smoothness of the linear transitions between the samples.
Unlike A3D~\cite{Ignatyev2025A3D}, our method is inference-based and does not use SDS, which makes it orders of magnitude faster, allows for reaching state-of-the-art visual quality results without artifacts, and provides a notable variety of samples avoiding mode collapse.
Unlike MatchDiffusion~\cite{Pardo2025MatchDiffusion}, our method optimizes for the plausibility of the linear transitions between samples, notably improving structural alignment. 
It also provides a flexible way to co-guide samples during the inference process.

In summary, we make the following contributions.
\begin{itemize}
    \item We propose a new algorithm for joint paired inference with rectified flow models, which is based on velocity-guided transport of a segment in latent space, and provide a theoretical justification for it.
    \item We derive an analytical way to convert the rectified model velocity field into a velocity transport field for joint segment transport.
    \item We demonstrate that the proposed algorithm, combined with models trained on structured latents, produces highly structurally aligned samples for three modalities: images, videos, and 3D models.
    \item We obtain results on par with state-of-the-art, specifically trained methods on images and 3D models, and show superior state-of-the-art results on videos.
\end{itemize}

\section{Related work}
\label{sec:related}
\subsection{Image Generation and Editing}
Image generation and editing are the two core problems in generative modeling and form the foundation for controllable video and 3D synthesis. 
The early approaches relied on Generative Adversarial Networks (GANs) \cite{goodfellow2020gan, kang2023scalinggan, huang2024deadgan} and Variational Autoencoders (VAEs) \cite{kingma2013vae, van2017vqvae}, which were later surpassed by higher-quality diffusion-based methods such as DDPM \cite{ho2020ddpm} and Stable Diffusion \cite{rombach2022sd}. 
The introduction of these models enabled scalable text-to-image generation \cite{song2020ddim, ramesh2022dalle2, saharia2022imagen}.

More recent progress in text-to-image modeling has been driven by the emergence of Rectified Flow \cite{liu2023rectifiedflow} and Diffusion Transformers (DiT) \cite{peebles2023dit}, which underpin several state-of-the-art systems, including FLUX.1 \cite{blackforestlabs_flux1dev_2024}, Stable Diffusion 3 \cite{esser2024sd3}, and Qwen-Image \cite{wu2025qwenimagetechnicalreport}. 
Their stable training and expressive conditioning have also made them effective when applied to image editing \cite{rout2025semanticedit, labs2025fluxkontext, qwen_image_edit_2025}.

\subsection{Video Generation and Editing}
Video-based models are still far behind image-based models due to computational requirements and data scarcity \cite{wan2025wanopenadvancedlargescale, hong2022cogvideolargescalepretrainingtexttovideo}. 
Existing video editing methods fall into two groups: training-free modifications of video generation models and methods that train specific editing networks.

Training-free approaches are easy to adapt to existing models.
A representative example is MatchDiffusion \cite{pardo2024matchdiffusiontrainingfreegenerationmatchcuts}, which relies on joint and disjoint diffusion stages, balancing between visual coherence and semantic divergence.

Methods requiring training allow more versatile and fine-grained edits. 
Among these methods, VACE \cite{jiang2025vaceallinonevideocreation} supports diverse conditions, including inpainting, depth, and motion preservation. 
Another work, LucyEdit \cite{decart2025lucyedit}, focuses on purely textual edits and special “trigger” words, allowing control of the granularity of edits.

\subsection{3D Generation and Editing}
3D object and scene synthesis has also advanced significantly.
Unlike 2D domains, where large-scale datasets enable highly generalizable models \cite{blackforestlabs_flux1dev_2024, wu2025qwenimagetechnicalreport}, limited 3D data availability has led researchers to exploit 2D priors from pretrained models—either through multi-view generation with subsequent reconstruction \cite{shi2023mvdream, li2023instant3d, go2025splatflow, szymanowicz2025bolt3d, huang2025mvadapter} or through score distillation from 2D diffusion models (\cite{poole2022dreamfusion, liang2023luciddreamerism, lukoianov2024sdi}).
More recently, diffusion and flow matching models that operate directly in a 3D latent space have become competitive \cite{xiang2024trellis, yang2025prometheus, lai2025hunyuan3d25, xiang2025native}. 
These methods fall into two main groups: those using structured voxel-based latents \cite{xiang2024trellis, wu2025unilat3d, yang2025prometheus, xiang2025native} and unstructured latents \cite{zhao2025hunyuan3d, lai2025hunyuan3d25, lai2025unleashing}. While unstructured latents are more computationally efficient, structured ones offer better interpretability—an advantage we build upon in our work. 
3D editing has followed a similar trajectory.
Early progress relied on the use of 2D priors: either via score distillation \cite{zhuang2024tip, li2024focaldreamer, zhuang2023dreameditor, chung2023luciddreamer} or via multi-view diffusion \cite{chen2024dge, mvedit2024, lee2025editsplat}.
Recently, methods operating directly in 3D latent space have enabled editing directly in 3D space \cite{xiang2024trellis, li2025voxhammer, ye2025nano3d}.
However, they still have limited generalization ability due to data scarcity, so 2D-based methods remain a strong baseline for this task. 

\subsection{Joint Generation of Consistent Objects}
Recent work has shown that generative models can jointly produce multiple objects with shared properties.
These properties may vary in nature: \cite{lee2023syncdiffusion, yeo2025stochsync} exploit synchronization mechanisms to produce multiple views of a scene that can subsequently be merged into a panorama, \cite{avrahami2023chosen} proposed a procedure to generate a set of images with consistent identity. 
The joint generation of geometrically aligned 3D assets remains more challenging, and the existing solution relies on an iterative, time-consuming SDS-based A3D algorithm \cite{Ignatyev2025A3D}. 
Its core idea is to enforce smooth transitions between generations. 
MatchDiffusion \cite{Pardo2025MatchDiffusion} proposes a training-free joint-generation approach that merges two trajectories before a threshold timestep, but we found this insufficient for reliable structural alignment, motivating our method.
\section{Preliminaries}
\label{sec:preliminaries}
\subsection{Flow Matching}
Recently, there emerged a tendency in the community to switch from the denoising diffusion models to simpler and more effective Rectified Flow~\cite{liu2023rectifiedflow} models. 
Given a distribution $X_0$ and a sample from it $x_0 \sim X_0$, the initial distribution is transformed into the "noised" version via linear interpolation with $\epsilon \sim \mathcal{N}(0, \bm{I})$, where $t \in [0, 1]$
\begin{equation}
    \label{equation:flow_matching}
    x_t = (1 - t)x_0 + t \epsilon .
\end{equation}
Then a transformer model is trained to directly regress the velocity field $v_{\theta}(x_t, t)$ by minimi
\begin{equation}
    \mathcal{L}(\theta)
=
\mathbb{E}_{t,\; x_0 \sim X_0,\; x_1 \sim X_1}
\left[
\left\| v - v_{\theta}(x_t, t) \right\|^{2}
\right].
\end{equation}
\section{Method}
\label{sec:method}
Multiple works \cite{Ignatyev2025A3D, berthelot2018understanding, karras2020analyzing} imply the importance of two requirements for learning the transitions between samples, which lead to the generation of the semantically aligned objects:
1) Transitions between the aligned objects should provide plausible and realistic samples. 
2) Transitions should be smooth (or have a bounded Lipschitz constant).
In this section, we analyze these requirements and suggest a principled algorithm for joint inference with rectified flow models, which can be combined with an arbitrary pre-trained model operating on the set of structured latents.

\begin{figure}[h]
    \centering
    \begin{subfigure}{0.45\textwidth}
        \centering
        \includegraphics[width=\textwidth]{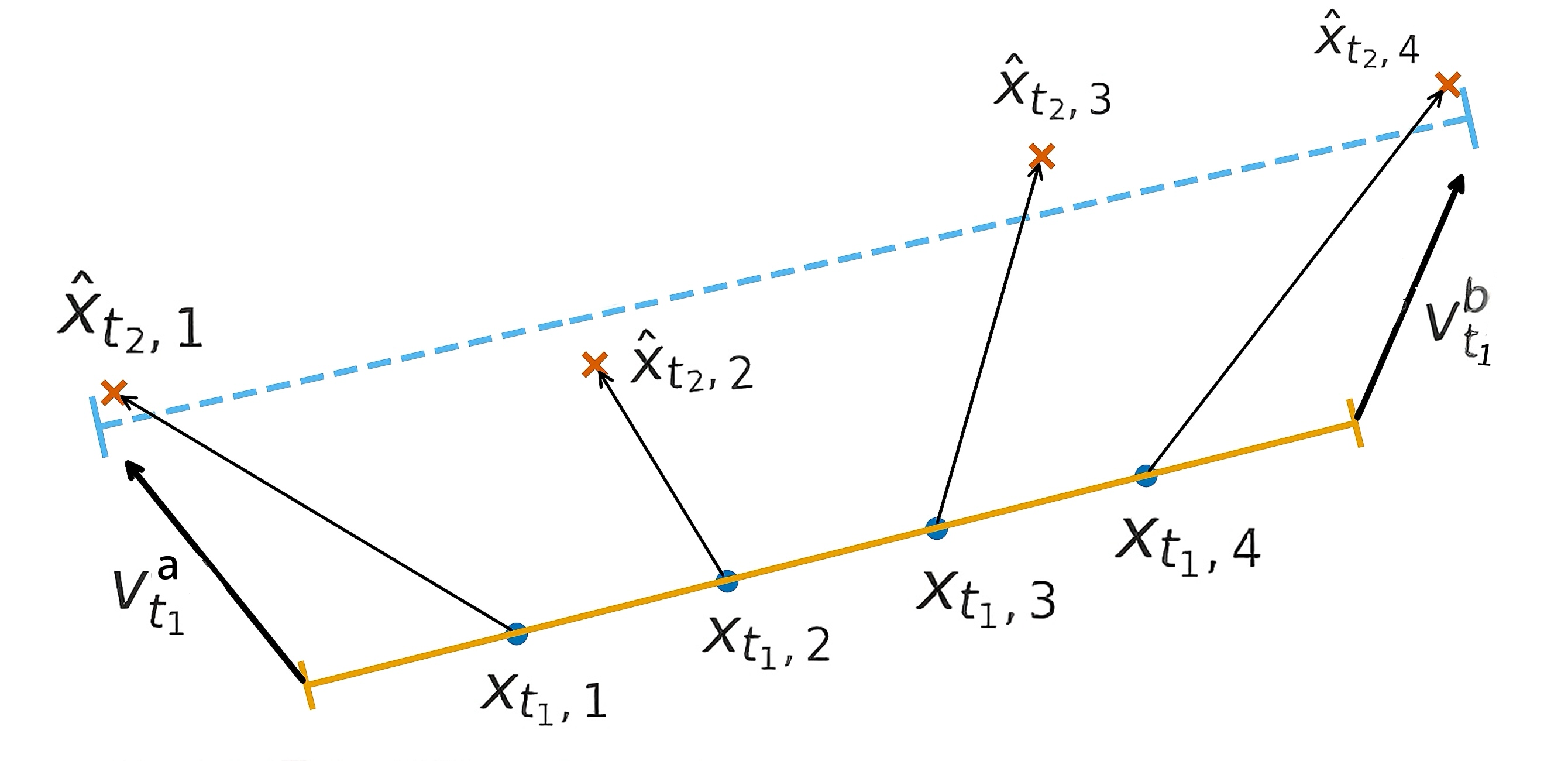}
        \caption{Joint Step}
        \label{fig:method_joint}
    \end{subfigure}
    \hfill
    \begin{subfigure}{0.45\textwidth}
        \centering
        \includegraphics[width=\textwidth]{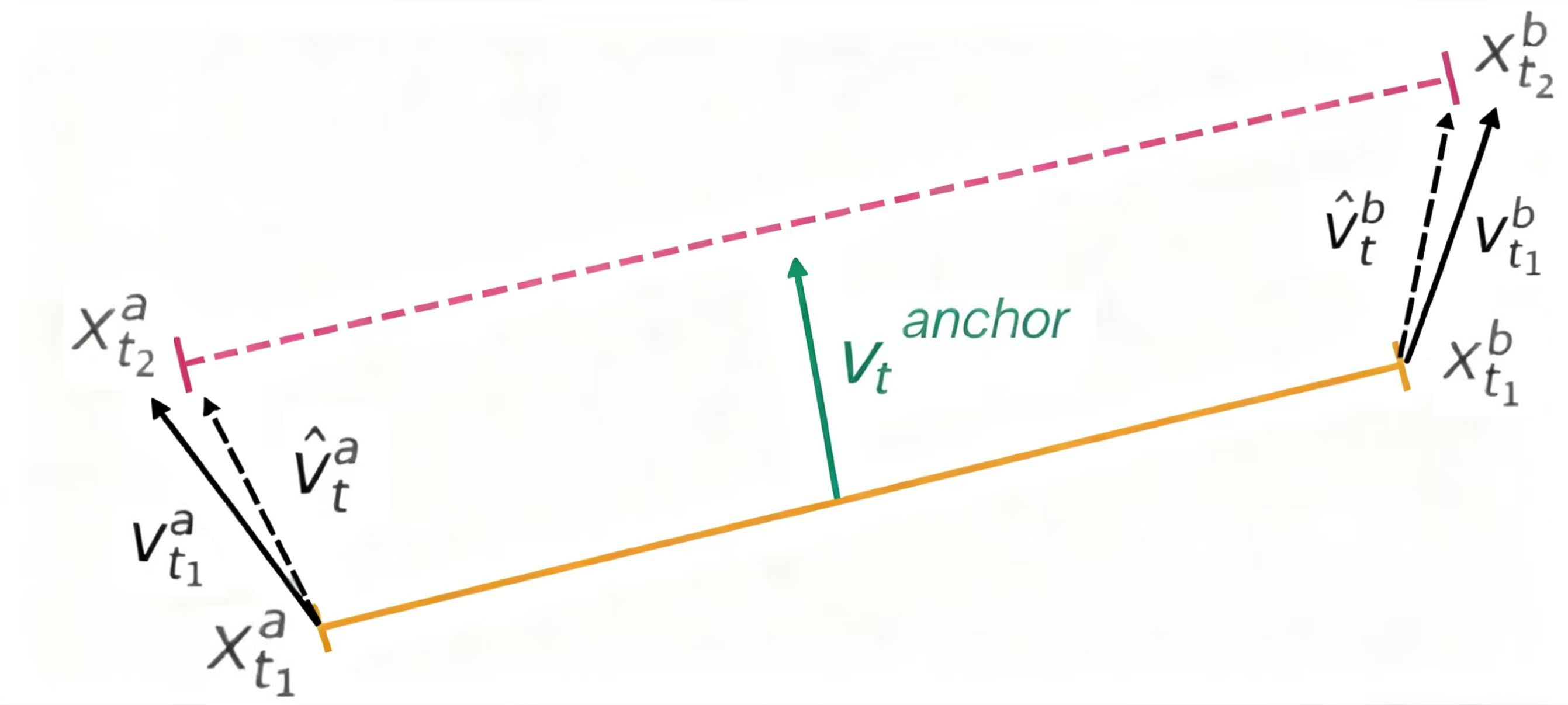}
        \caption{Smoothness Correction}
        \label{fig:method_smoothness}
    \end{subfigure}
    \caption{Method}
    \label{fig:complete}
\end{figure}

\subsection{Joint Inference with Rectified Flow Models}
\label{sec:joint_inference}
Flow-matching models use time discretization to approximate trajectories along the velocity vector field $v_{\Theta}(x_t, t, c)$, which is parameterized by a neural network.
Given a text embedding $c$ and starting with a sample $x \sim \mathcal{N}(0, \bm{I})$ taken from a Gaussian noise distribution, the sample~$x_{t_{1}}$ at time step~\(t_1\) can be used to calculate $x_{t_{2}}$ (where $t_1>t_2$) with the following update rule:
\begin{equation}
    \label{eq:rectified_inference}
    x_{t_2} = x_{t_1} + (t_2 - t_1) v_{\Theta}(x_{t_1}, t_1, c).
\end{equation}
Let $c^a$, $c^b$ be two text embeddings and $x^a$, $x^b$ be the sample variables corresponding to a pair of objects \(a\) and \(b\).
We initialize $x^a$ and $x^b$ with the same value.
Instead of transporting them independently, we want to do it jointly to improve the plausibility of the transitions between them.
To achieve this, we consider transporting a distribution of samples on the line segment $[x^a,x^b] \;=\; \{ (1-\alpha)\,x^a + \alpha\,x^b \mid \alpha \in [0,1] \}$ defined by some density function~\(p(\alpha)\).
Thus, we shift from transporting individual samples to transporting probability distributions, while preserving the linear structure of these distributions.


To define the update rule for joint transport, we represent the samples distributed on the line segment with the density~\(p(\alpha)\) by a set of weighted samples evenly distributed across the segment $\{\, x_{t, i} = (1-\alpha_i)\,x^a_t + \alpha_i\,x^b_t \;\mid\; i = 1,\dots,k \,\}$ with the respective weighting factors~\(p(\alpha_i)\).
Given the segment $[x^a_{t_1},x^b_{t_1}]$ we update it to the segment $[x^a_{t_2},x^b_{t_2}]$ in two steps. 
First, we update each point $x_{t_1,i}$ to $\hat{x}_{t_2, i}$ individually using the rule in Equation~\ref{eq:rectified_inference} and the interpolated text embedding $c_i=(1-\alpha_i)c^a + \alpha_i c^b$. 
While the points $\{\, x_{t_1, i} \;\mid\; i = 1,\dots,k \,\}$ by definition always lie on a single line in high-dimensional space there is, generally speaking, no such guarantee for the updated points $\{\, \hat{x}_{t_2, i} \;\mid\; i = 1,\dots,k \,\}$. 
That is why we restore the linear structure of the distribution by solving a linear regression problem in Equation~\ref{eq:linear_regression}, minimizing $\mathcal{L}$ \text{w.r.t.} $x^a_{t_2}$ and $x^b_{t_2}$.
\begin{equation}
\label{eq:linear_regression}
\begin{split}  
    \mathcal{L}(x^a_{t_2}, x^b_{t_2})  = \sum_{i=1}^{k} p(\alpha_i)\| x_{t_2, i} - \hat{x}_{t_2, i} \|_2,\\
    \hat{x}_{t_2,i} = x_{t_1,i} + (t_2 - t_1)\, v_{\Theta}\bigl(x_{t_1,i}, t_1, c_i\bigr),\\
    x_{t_2,i} = (1-\alpha_i)\,x^a_{t_2} + \alpha_i\,x^b_{t_2}.
\end{split}
\end{equation}
This regression problem has an explicit solution. 
We define:

Then the optimal endpoints are:
\begin{equation}
x_{t_2}^a = \frac{c_{11} d_0 - c_{01} d_1}{\Delta}, \quad
x_{t_2}^b = \frac{c_{00} d_1 - c_{01} d_0}{\Delta}.
\end{equation}
The described approach enables a rapid update of the distribution parameters, thereby avoiding the slow gradient-based optimization.
Finally, we can define the velocities of the parameters of the probability distribution (boundary points of the segment $x^a_t$ and $x^b_t$): 
The scheme of the joint update is shown in Figure~\ref{fig:method_joint}.
\begin{equation}
\label{eq:parameter_velocity}
v^a_{t_1} = \frac{x_{t_2}^a - x_{t_1}^a}{t_2 - t_1}, \quad v^b_{t_1} = \frac{x_{t_2}^b - x_{t_1}^b}{t_2 - t_1}.
\end{equation}
In the presented way, we transform the transport velocity field for samples into the joint transport velocity field for probability distributions supported by the segments in the sample space.
In practice, we found it important to use the density $p(\alpha)$ centered around the midpoint during the early iterations and gradually shifted towards a uniform distribution.

\subsection{Smoothness regularization}
The second aspect is the need for smooth transitions between the two objects.
For linear transitions, the "speed" of transition does not depend on the point and is always the same: $\frac{dx_{t}}{d\alpha} = \frac{d}{d\alpha}((1-\alpha)\,x^a_t + \alpha x^b_t)=x^b_t - x^a_t$.
Thus, regularizing the "speed" of transition is the same as regularizing the norm $||x_t^b - x_t^a||_2$ of the segment $[x_t^a, x_t^b]$.
Since we initialize $x^a$ and $x^b$ with the same value, the segment norm is zero in the beginning, and over time it diverges, resulting in different, but structurally aligned samples.

The derivative of the \(L^2\)~norm of the segment can be written as
\begin{equation}
\label{eq:segment_dynamics}
\frac{d||x^b(t) - x^a(t)||_2}{dt} = \frac{ \langle v^b(t) -v^a(t), x^b(t) - x^a(t) \rangle } {||x^b(t) - x^a(t)||_2},
\end{equation}
where \(\langle \cdot, \cdot \rangle\) is the dot product.
We observe that in the cases where we see the severe misalignment between the samples, it usually correlates with the rapid growth of the segment norm during the early iterations of the inference. 
We can see that this derivative depends on the difference between the velocities of the endpoints of the segment $v^b-v^a$. 
Thus, minimizing the derivative can be achieved by minimizing this expression or making the velocities of the distribution parameters closer to each other. 
One particular way to achieve this is to correct the velocities by pulling them closer to some specific \textit{anchor} velocity $v^{anchor}$ (Equation~\ref{eq:anchor})
\begin{equation}
\label{eq:anchor}
\hat{v}^a_t=w_tv^{anchor}_t + (1-w_t)v_t^a, \quad \hat{v}^b_t=w_tv^{anchor}_t + (1-w_t)v_t^b.
\end{equation}
The choice of $v^{anchor}$ is important - the anchor velocity vector should point at the "denoising" direction for all the points on the segment in order not to break the noise schedule of the samples.
One possible solution for this task would be to choose $v_t^{anchor}=\frac{v_t^a + v_t^b}{2}$. 
Though our experiments show that this solution is suboptimal, probably because these two endpoint velocities often have conflicting directions.
Our suggested solution is to choose $v^{anchor}$ as a predicted velocity (Equation~\ref{eq:our_anchor}) for the midpoint of the segment 
\begin{equation}
\label{eq:our_anchor}
v_t^{anchor} = v_{\Theta}(\frac{x_t^a + x_t^b}{2}, t, \frac{c^a+c^b}{2}).
\end{equation}
Note that such choice of $v^{anchor}$ is essentially inseparable from the approach presented in Section~\ref{sec:joint_inference} serving as a "correction" to the formulated velocity field, because for the proper inference $\frac{x_t^a + x_t^b}{2}$ should be a plausible sample from the noised distribution at the timestep $t$ which requires taking into account the velocities for the intermediate points of the segment.
The scheme for the smoothness correction mechanism is shown in Figure~\ref{fig:method_smoothness}.
We observe that choosing the moderate schedule for the values $w_t$ with the emphasis on early iterations does not lead to the degradation of the final samples compared to the "base" rectified flow method. 

\begin{figure*}
    \centering
    \begin{subfigure}[t]{0.24\linewidth}
      \centering
      \includegraphics[width=\linewidth]{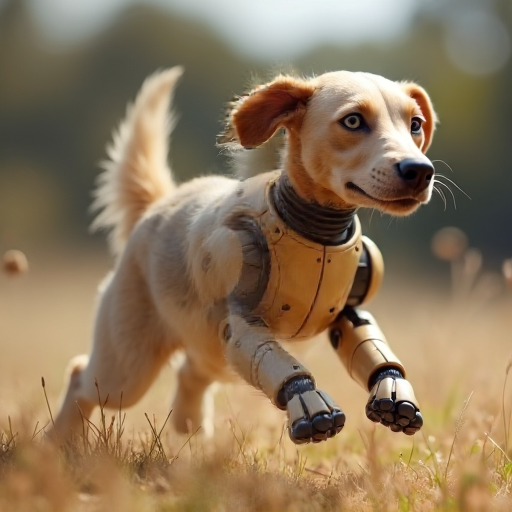}
      \caption{dog, robot}
    \end{subfigure}\hfill
    \begin{subfigure}[t]{0.24\linewidth}
      \centering
      \includegraphics[width=\linewidth]{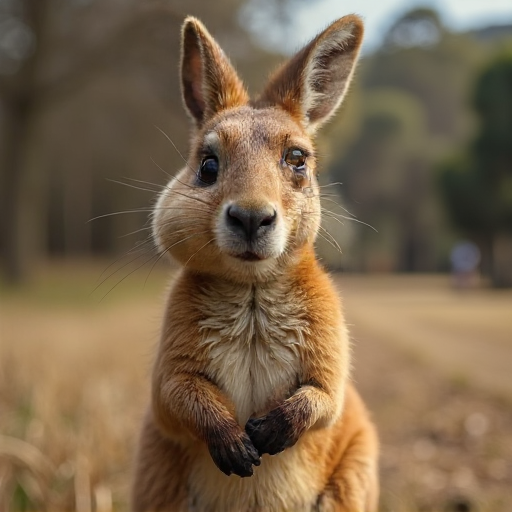}
      \caption{gopher, kangaroo}
    \end{subfigure}\hfill
    \begin{subfigure}[t]{0.24\linewidth}
      \centering
      \includegraphics[width=\linewidth]{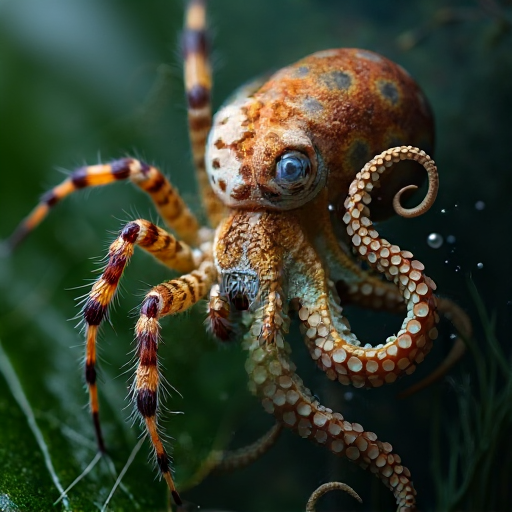}
      \caption{spider, octopus}
    \end{subfigure}\hfill
    \begin{subfigure}[t]{0.24\linewidth}
      \centering
      \includegraphics[width=\linewidth]{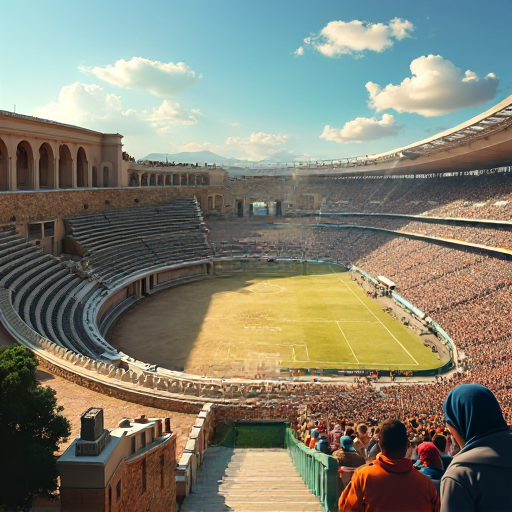}
      \caption{amphitheatre, stadium}
    \end{subfigure}
    \caption{Visualization of geometry preservation between two generated images. For each example, two images are blended into one with a blending coefficient $\alpha(column)$ that depends on the column index, increasing from 0 (left side) to 1 (right side). With such blending, we show that not only geometry is preserved, but also that smooth transitions between two generations are enabled.}
    \label{fig:alpha_blending}
\end{figure*}

\section{Experiments}
\label{sec:experiments}
We show the universality of our method for joint aligned generation by applying it to three major domains for generative modeling: images, 3D, and video. 
We select three state-of-the-art rectified flow pre-trained models, which operate on structured latents: voxels, pixels, and video frames.
We modify the inference loops of these three models in a self-contained way, which requires only minor differences between the models.
We do not change model weights or any other parts of the pipeline.

We use two types of methods as competitors i) joint generation methods, which produce paired output at single inference (A3D~\cite{Ignatyev2025A3D}, MatchDiffusion~\cite{Pardo2025MatchDiffusion}) ii) editing-based methods which take an independently generated sample from the first prompt \textit{source} and complement the pair by editing it with another prompt from the pair (RF-Inversion~\cite{rout2025semanticedit}, VACE~\cite{vace}, and MVEdit~\cite{mvedit2024}).

\textbf{Metrics:} We use multiple metrics to evaluate the degree of structural alignment between the samples and consistency between the samples and the corresponding textual descriptions. 

\textit{Modality agnostic metrics} are based on the evaluation of the sample "projections" to the 2D image space (multi-view images, video frames).
\textbf{DIFT} alignment \textbf{score} was introduced in A3D~\cite{Ignatyev2025A3D} as a way to measure structural similarity between two salient objects on a pair of images.
It works by building a dense grid of 2D points on the source image and finding the corresponding point for each point in the grid using the DIFT~\cite{tang2023emergent} method.
The pairwise distance between the points is averaged. Finally, the metric is calculated in the reverse direction and averaged one more time.
When calculating the DIFT Score, we use the Grounded SAM~\cite{ren2024grounded} segmentation model to focus on the foreground object to avoid matching the background.
In the recent work SPIE~\cite{Benarous2025SPIE} $L_1$ distance (or equivalently, MAE mean absolute error) between the \textbf{depth} maps extracted from the image pair was shown to be an effective proxy metric for structural alignment.
We extract depth maps for evaluation using the Depth Anything V2 model~\cite{depth_anything_v2}.
We also employ the widely-used \textbf{CLIP Score} to estimate how well each prompt fits the corresponding 2D image by calculating the similarity between the prompt embedding and image embedding.
On par with CLIP, we use specific subcriteria from the MLLM\cite{huang2025t2icompbench++} method to evaluate image-prompt alignment.

\textit{Video metrics} are specifically designed to evaluate the qualities of videos.
The temporal consistency of the videos is evaluated by calculating the similarity between the \textbf{DINO} features of neighboring frames for the edited video. 
To assess the overall quality of the videos, we use a \textbf{VLM} score~\cite{ju2025editverseunifyingimagevideo} to evaluate the overall quality, prompt alignment, and preservation of the details of the source video.

\textit{3D metrics}. 
We use \textbf{GPTEval}~\cite{Wu2024GPTEval3D}, a VLM-based method to evaluate the quality of 3D objects, including geometry, texture, and prompt consistency.

\subsection{Images}
We build our joint image generation pipeline on top of the widely used FLUX.1-dev~\cite{blackforestlabs_flux1dev_2024}, which is known for the excellent quality of text-to-image generation.
For the consistency with 3D experiments, we use the set of object pairs from A3D~\cite{Ignatyev2025A3D} with short prompts.
We compare with two editing-based methods, RF-Inversion~\cite{rout2025semanticedit}, which is a flow-inversion training-free method, and instruction-based Qwen-Image-Edit \cite{wu2025qwenimagetechnicalreport,qwen_image_edit_2025}.
To obtain \textit{source} editing-free samples, we use FLUX.1-dev model.
We rewrite prompts into instructions for Qwen-Image-Edit, asking it to change the content of the source image so that the geometry and background are preserved. 
For image pair evaluation, we use DIFT Score, Depth Structural Score, CLIP Score, and MLLM Score.
\begin{table}[t]
    \centering
    \caption{2D Metrics.}
    \includegraphics[max height=7\baselineskip, max width=0.4\textwidth]{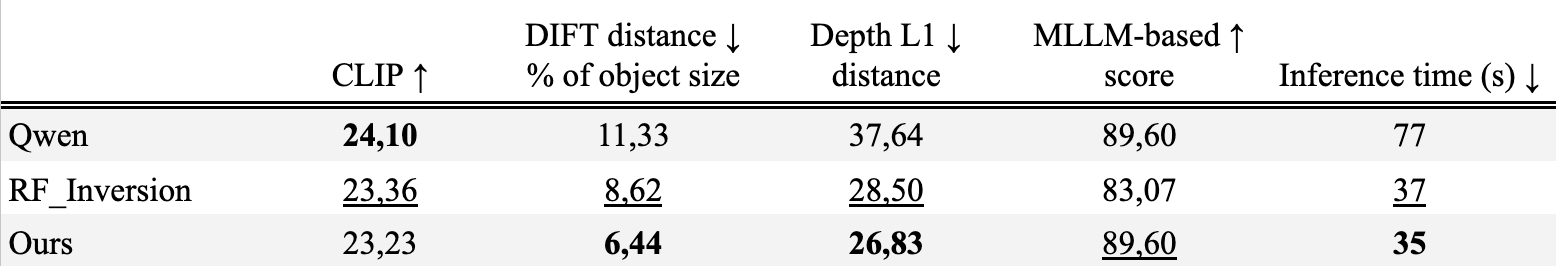}
    \label{tab:2d_metrics}
\end{table}
The quantitative results are presented in the Table~\ref{tab:2d_metrics}.
Our method notably improves over RF-Inversion in terms of structural alignment metrics.
While Qwen demonstrates a superior understanding of user instructions, it lags far behind in structural alignment score, which makes it unsuitable for this particular problem.

Our method can be used to seamlessly combine parts of the samples from parallel domains. 
We demonstrate it by changing the $\alpha$ blending coefficient across the horizontal axis for the pairs of images in Figure~\ref{fig:alpha_blending}.

\subsection{3D Shapes}
For joint 3D object generation, we build on two pipelines: Trellis~\cite{xiang2024trellis}, a text-to-3D method, and its improved image-to-3D version, Trellis.2~\cite{xiang2025native}.
The Trellis pipeline consists of two parts: \textit{Structure Generation} operating on dense voxels and \textit{Structured Latents Generation} working with a sparse latent representation. 
To generate geometrically aligned 3D models, we modify the inference loop for the rectified flow model in the first part of the pipeline, leaving everything else intact.

\begin{figure*}[t]
    \centering
    \caption{3D Alignment Samples}
    \includegraphics[width=\textwidth]{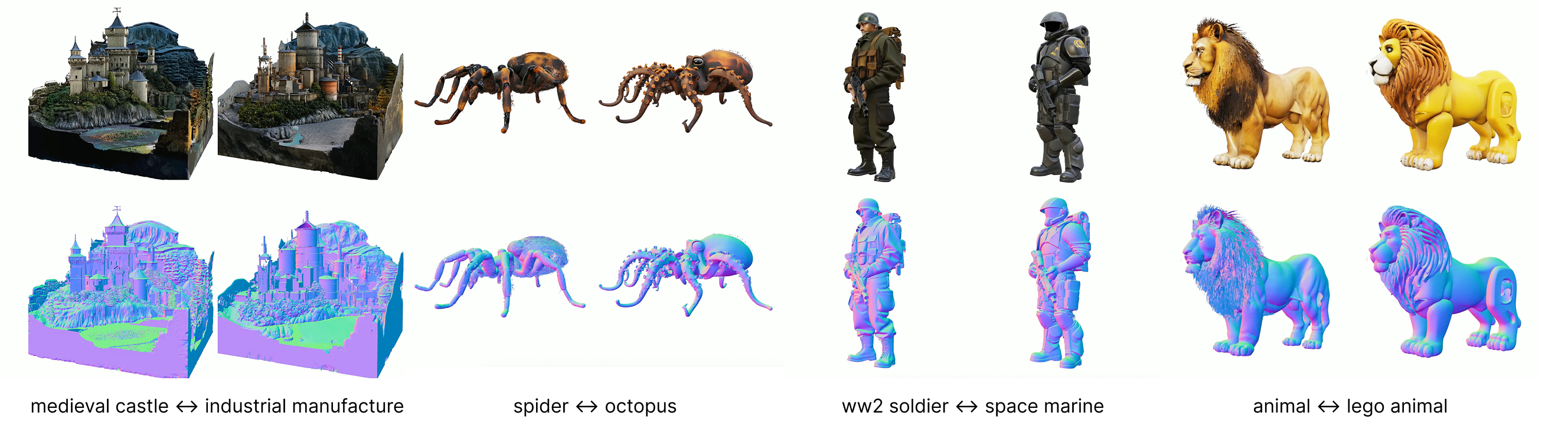}
    \label{fig:3d_alignment}
\end{figure*}

\begin{table}[t]
    \centering
    \caption{GPTEval 3D Metrics, \% of comparisons where our method based on Trellis.2 is preferred over competitors and over our method based on Trellis.1.}
    \includegraphics[max height=7\baselineskip, max width=0.49\textwidth]{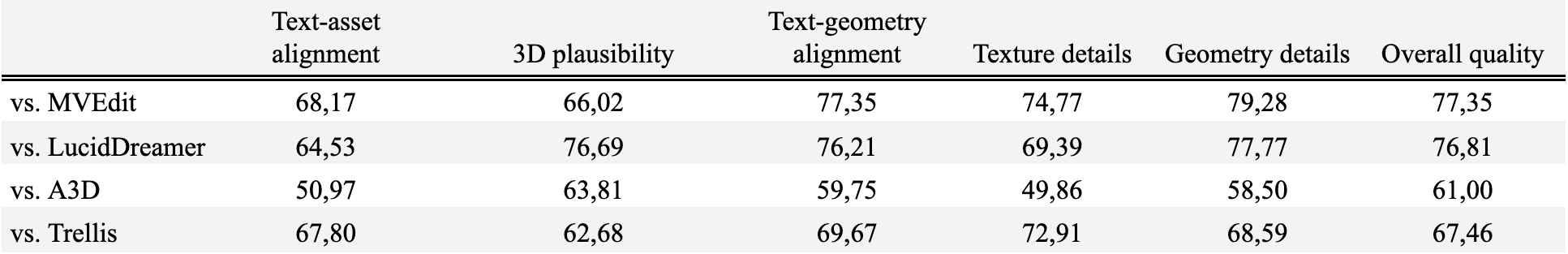}
    \label{tab:gpt_eval}
\end{table}
\begin{table}[b]
    \centering
    \caption{Other 3D Metrics.}
    \includegraphics[max height=7\baselineskip, max width=0.47\textwidth]{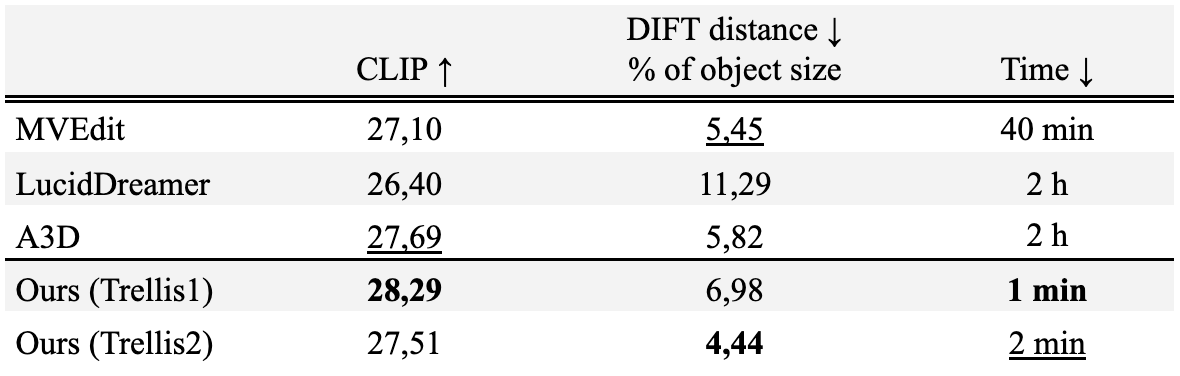}
    \label{tab:3d_metrics}
\end{table}
Because Trellis was trained primarily on detailed text descriptions, the short prompts used in the A3D~\cite{Ignatyev2025A3D} evaluation introduce a distribution shift.
Therefore, we rewrite the short prompts into more detailed descriptions using GPT-5.
With Trellis.2, the difference in our approach  is that the interpolated input conditions are image tokens rather than text embeddings.
To generate aligned 3D objects from aligned images, we use image pairs produced by our image joint generation method, built on top of the Flux model.
Trellis.2 and Flux joint generation methods together result in text-to-3D pipeline. 
We compare our approach with editing-based methods (MVEdit~\cite{mvedit2024}, LucidDreamer~\cite{liang2023luciddreamerism}) and with the joint generation method A3D.
The source 3D models for the editing-based methods are obtained using the generative pipelines associated with each method.
For evaluation, we use the DIFT score, GPTEval score, and CLIP score.
The results are shown in Tables~\ref{tab:gpt_eval}  and~\ref{tab:3d_metrics}.
Our method shows the best CLIP score for text-to-3D semantic alignment.
Regarding DIFT structural alignment, our alignment correction combined with the Trellis.1 pipeline yields strong results, albeit slightly lagging behind the exceptionally strong A3D and MVEdit scores because of the limited generalization ability of the backbone model. 
When incorporated into the Trellis.2 pipeline, our method shows strongest alignment results with decent generation quality. 
On GPTEval, it confidently improves over the competitors' results.
Another important quality of our method is its speed. 
It provides an order-of-magnitude speedup compared with A3D and is significantly faster than MVEdit.
Qualitative results are shown in the Figure~\ref{fig:3d_alignment}, demonstrating the high degree of geometric alignment.

\subsection{Video}
\begin{figure}
  \centering
  \begin{minipage}[t]{0.4\textwidth} 
    \centering
    \begin{subfigure}[t]{0.49\linewidth}
      \centering
      \includegraphics[width=\linewidth]{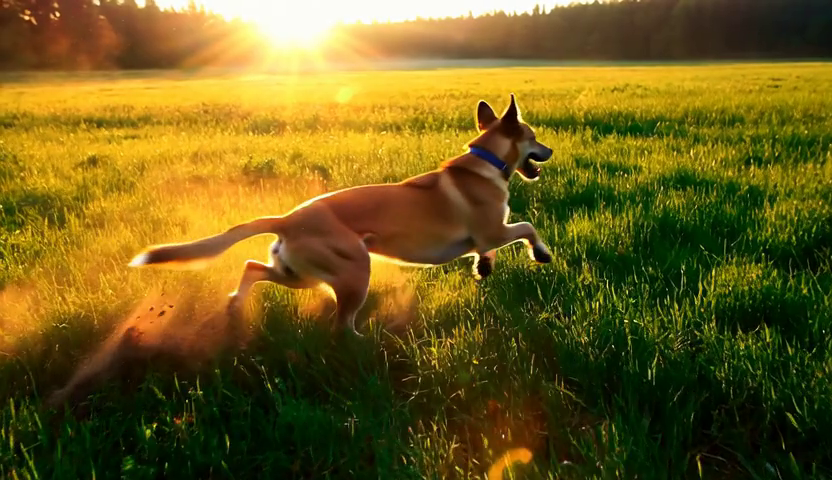}
      \caption{Our method's source video frame}
    \end{subfigure}\hfill
    \begin{subfigure}[t]{0.49\linewidth}
      \centering
      \includegraphics[width=\linewidth]{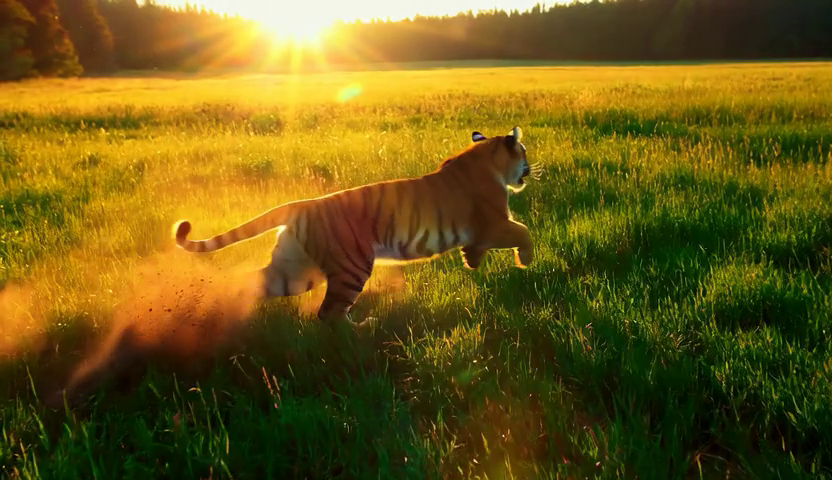}
      \caption{Our method's target video frame}
    \end{subfigure}

    \vskip 4pt 

    \begin{subfigure}[t]{\linewidth}
      \centering
      \includegraphics[width=\linewidth]{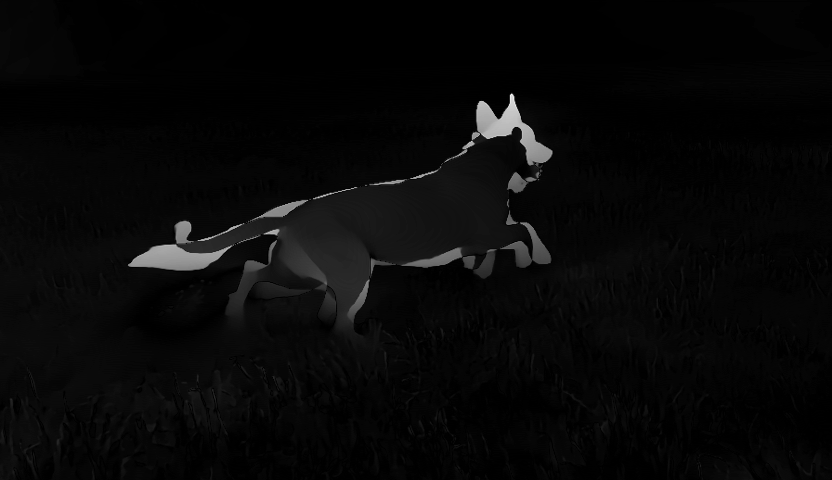}
      \caption{Depth difference between our method's source and target video frames}
    \end{subfigure}
  \end{minipage}
  \hfill
  \begin{minipage}[t]{0.4\textwidth} 
    \centering
    \begin{subfigure}[t]{0.49\linewidth}
      \centering
      \includegraphics[width=\linewidth]{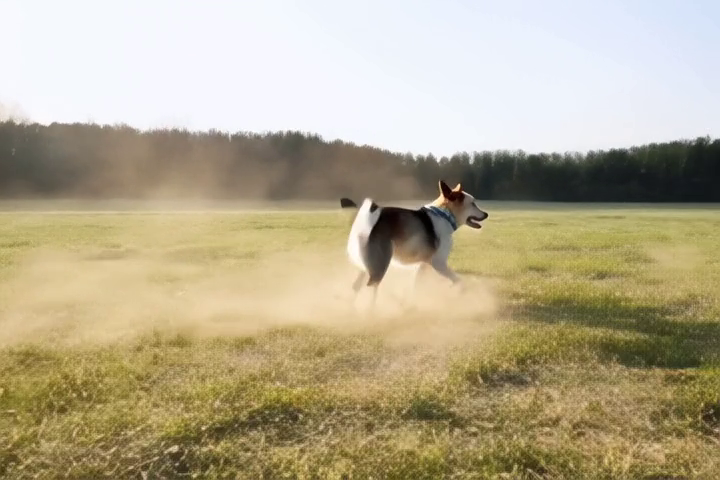}
      \caption{MatchDiffusion's source video frame}
    \end{subfigure}\hfill
    \begin{subfigure}[t]{0.49\linewidth}
      \centering
      \includegraphics[width=\linewidth]{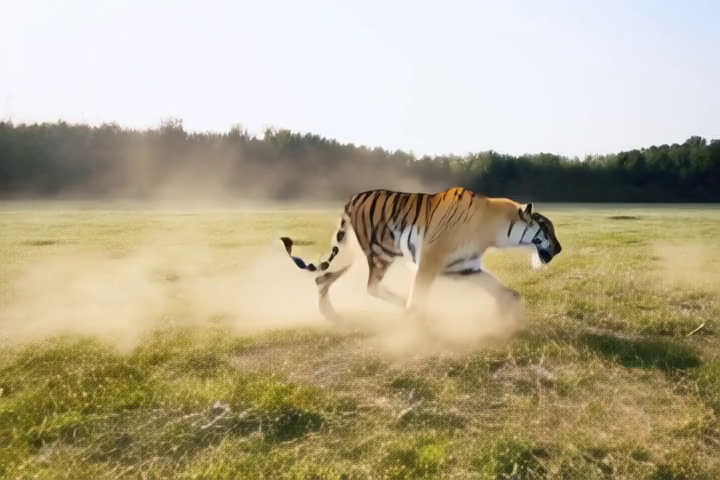}
      \caption{MatchDiffusion's target video frame}
    \end{subfigure}

    \vskip 4pt

    \begin{subfigure}[t]{\linewidth}
      \centering
      \includegraphics[width=\linewidth]{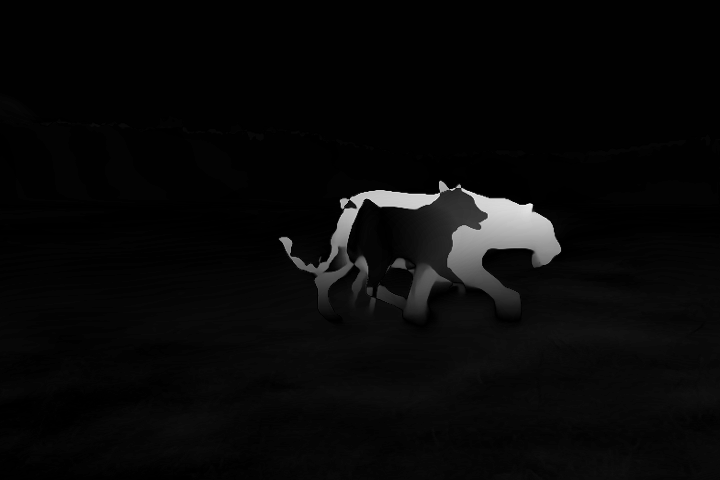}
      \caption{Depth difference between MatchDiffusion's source and target video frames}
    \end{subfigure}
  \end{minipage}

  \caption{Visualization of geometry preservation between source and target video frames. Larger white areas mean higher difference between corresponding pixels.}
  \label{fig:depth_difference}
\end{figure}

For video experiments, we modify the WAN 2.1~\cite{wan2025open} rectified flow model with our method.
Due to the dynamic nature of video samples, unlike the static 3D objects and images, we compose a novel set of scenes for evaluation. We aim to cover diverse and complex scenarios, including animals in motion, cities, and human activities.
We use two editing-based competitors, LucyEdit~\cite{decart2025lucyedit} and VACE~\cite{vace} (using depth-conditioned ControlNet).
For editing-based models, we use WAN-generated videos as a source video.
We also compare our setup with the joint generation method MatchDiffusion~\cite{Pardo2025MatchDiffusion}.
Depth-conditioned VACE tends to preserve the overall layout but substantially alters other aspects, including poses, motion, objects, and background.
On the other hand, LucyEdit can produce precise manipulations but is observed to work well only for humans and some animals, making nonsensical edits for the majority of the examples.
Quantitative results are shown in Table~\ref{tab:video_metrics}.
Our method achieves the highest DINO score, indicating the best self-consistency.
While LucyEdit formally shows lower depth MAE than our method, the reason for this is quite trivial - LucyEdit fails to provide necessary edits, leaving inputs intact. 
This is reflected in its very low VLM score, indicating poor alignment with the text.
Despite the depth-conditioned setup, VACE has the worst depth MAE, likely due to distribution shift induced by changing the prompt relative to the first frame.
On the other hand, MatchDiffusion shows good results both for VLM and DINO metrics but lags behind in terms of depth alignment, highlighting the need for a more principled approach.
Figure~\ref{fig:depth_difference} visualizes depth absolute error for MatchDiffusion and our method.
Both methods show small depth differences in the background, but MatchDiffusion produces much stronger errors around foreground object boundaries, indicating pose misalignment.
User-study results are reported in Table~\ref{tab:user_study}. 
The user study favors our method over VACE and LucyEdit across the main criteria.
It also supports our hypothesis of superior structural alignment, where our method shows a clear lead over MatchDiffusion.

\begin{table}[t]
    \centering
    \caption{Video Metrics.}
    \includegraphics[max height=7\baselineskip, max width=0.5\textwidth]{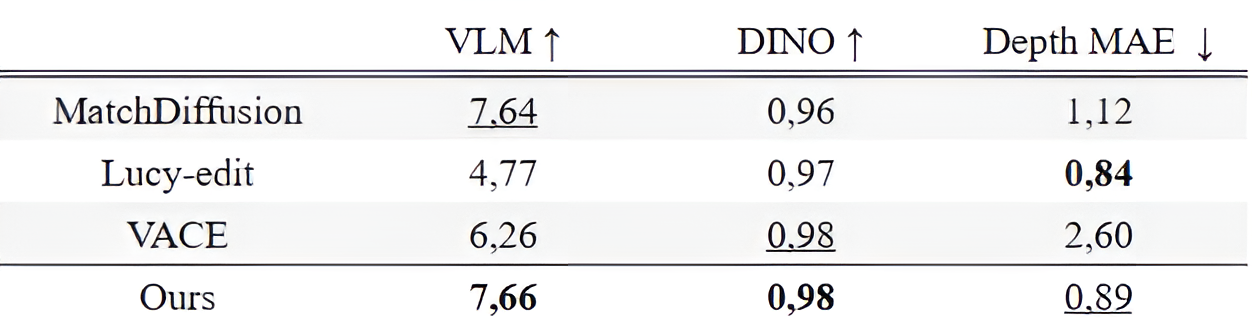}
    \label{tab:video_metrics}
\end{table}
\begin{table}[t]
    \centering
    \caption{User study.}
    \includegraphics[max height=7\baselineskip, max width=0.5\textwidth]{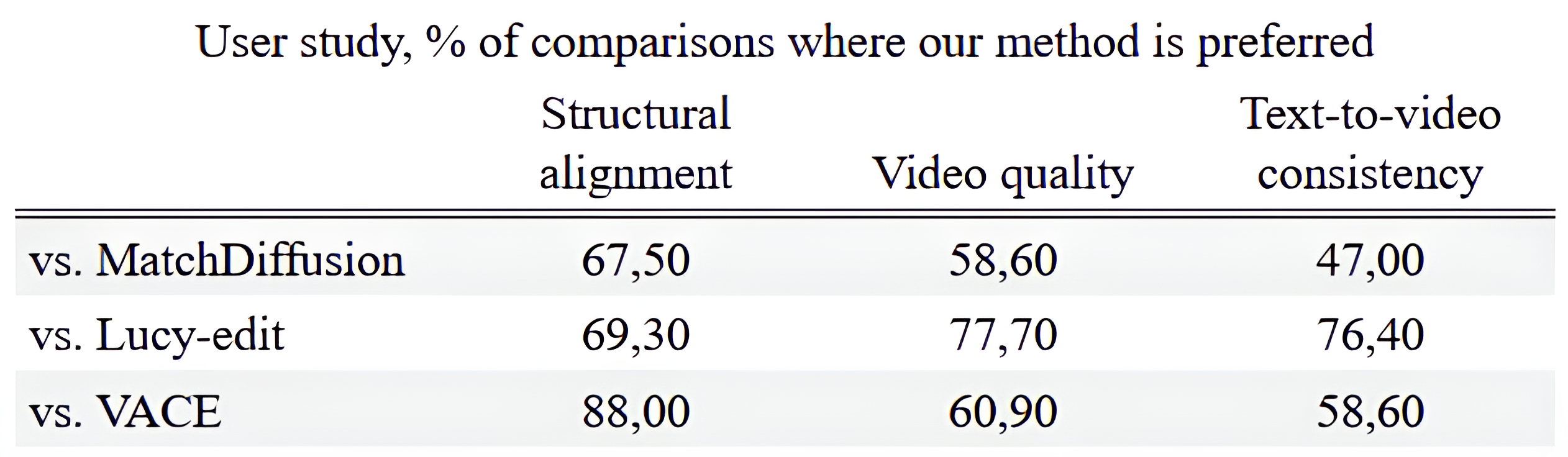}
    \label{tab:user_study}
\end{table}
\section{Ablation}
\label{sec:ablation}
\begin{figure}
    \centering
    \begin{subfigure}[t]{0.24\linewidth}
      \centering
      \includegraphics[width=\linewidth]{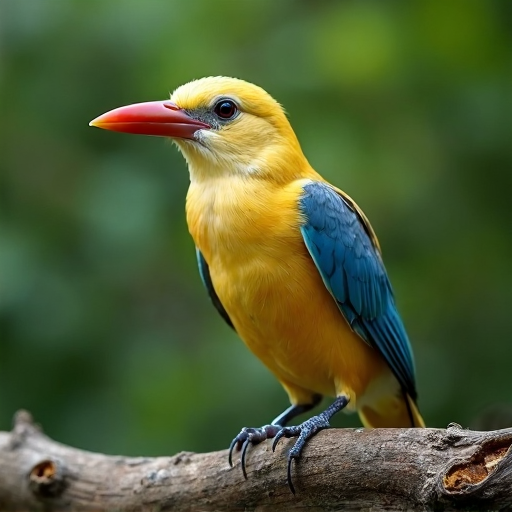}
    \end{subfigure}\hfill
    \begin{subfigure}[t]{0.24\linewidth}
      \centering
      \includegraphics[width=\linewidth]{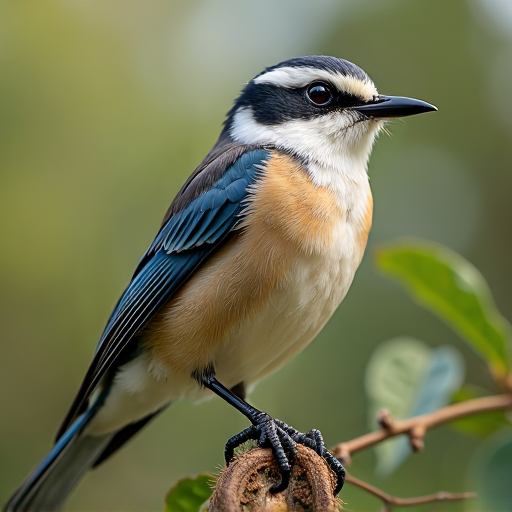}
    \end{subfigure}\hfill
    \begin{subfigure}[t]{0.24\linewidth}
      \centering
      \includegraphics[width=\linewidth]{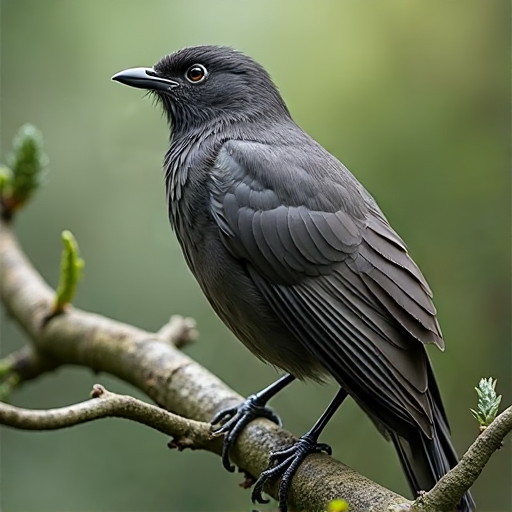}
    \end{subfigure}\hfill
    \begin{subfigure}[t]{0.24\linewidth}
      \centering
      \includegraphics[width=\linewidth]{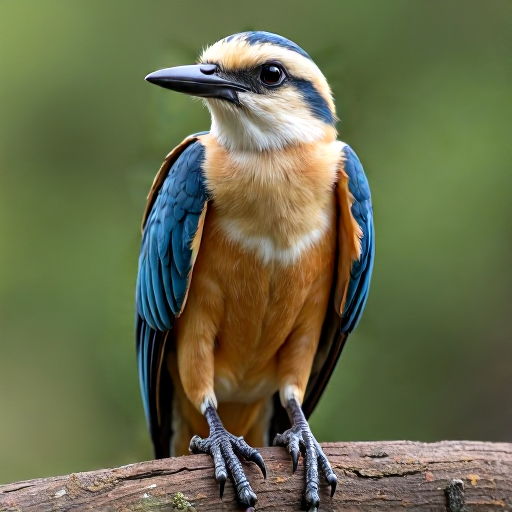}
    \end{subfigure}
    \vskip 4pt 
    \centering
    \begin{subfigure}[t]{0.24\linewidth}
      \centering
      \includegraphics[width=\linewidth]{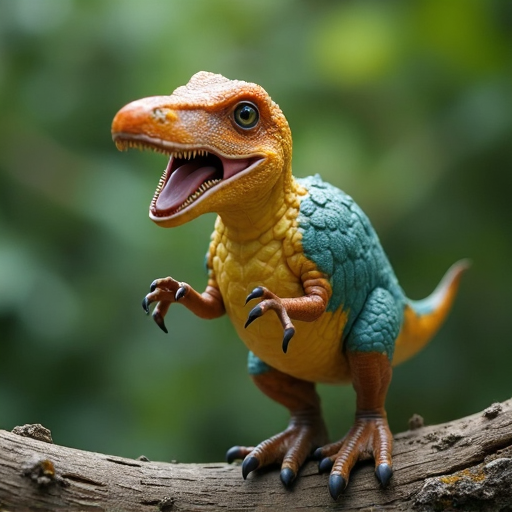}
      \caption{Setup A (ours)}
    \end{subfigure}\hfill
    \begin{subfigure}[t]{0.24\linewidth}
      \centering
      \includegraphics[width=\linewidth]{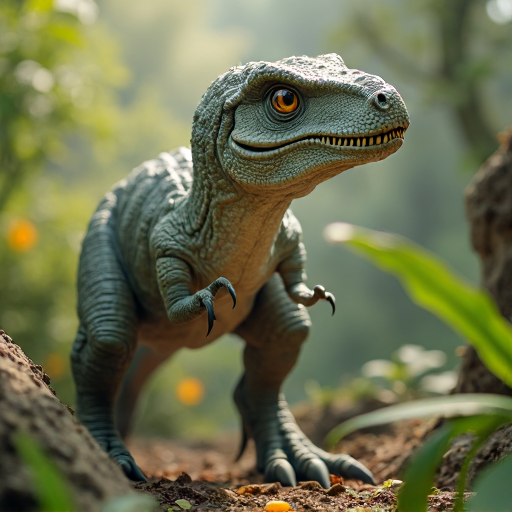}
      \caption{Setup B}
    \end{subfigure}\hfill
    \begin{subfigure}[t]{0.24\linewidth}
      \centering
      \includegraphics[width=\linewidth]{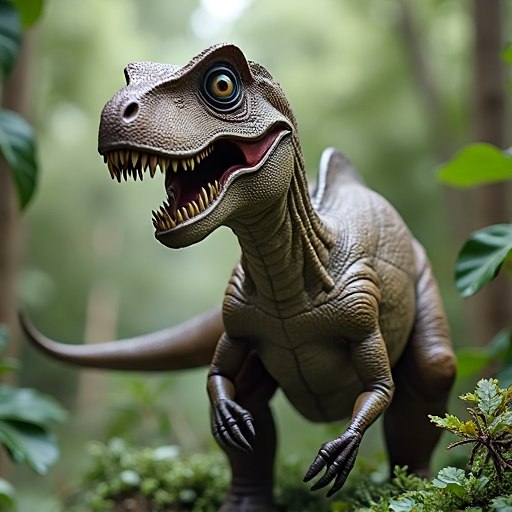}
      \caption{Setup C}
    \end{subfigure}\hfill
    \begin{subfigure}[t]{0.24\linewidth}
      \centering
      \includegraphics[width=\linewidth]{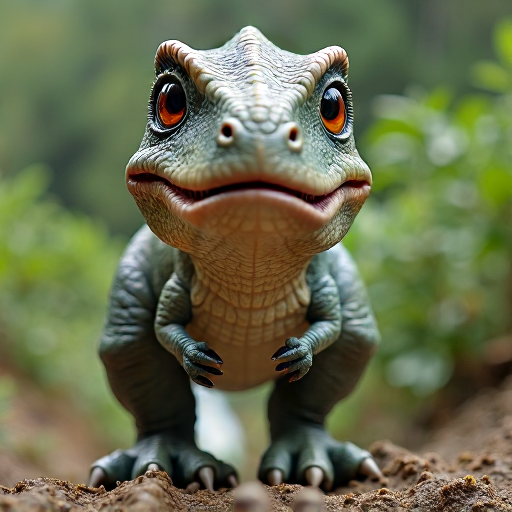}
      \caption{Setup D}
    \end{subfigure}
    \caption{Visualization of the impact of different components of our method. We can see that each component contributes to the results and that the results degrade as we gradually remove components from our algorithm.}
    \label{fig:ablation_study}
\end{figure}

To validate the design choices of our algorithm, we progressively remove components from our algorithm. 
After a series of such simplifications, our method effectively reduces to MatchDiffusion~\cite{Pardo2025MatchDiffusion}, yielding a sequence of controlled comparisons that isolates the contribution of each component. 
We start from the setup (A), which is our full pipeline described in Section~\ref{sec:method}.
Setup (B) is obtained by switching $v_t^{anchor}=\frac{v_t^a + v_t^b}{2}$ from $v_{\Theta}(\frac{x_t^a + x_t^b}{2}, t)$.
Setup (C) further removes sampling of intermediate points $x_{t, i}$.
Without intermediate points, the method no longer enforces plausible transitions between samples, which is central to our approach.
When we move to setup (D), the only restriction synchronizing the movement of $x_t^a$ and $x_t^b$ left in place is the anchor velocity $v_t^{anchor}=\frac{v_t^a + v_t^b}{2}$.
Instead of using the smooth schedule for weight coefficients, we simplify it to a hard cutoff 
which makes the algorithm effectively equivalent to the MatchDiffusion~\cite{Pardo2025MatchDiffusion} baseline. 
This step deprives the algorithm of a flexible co-guidance mechanism for the early stages of inference and removes any form of synchronization for the late stages.
\begin{table}[t]
\centering
\begin{subfigure}{0.48\columnwidth}  
    \centering
    \includegraphics[width=\linewidth]{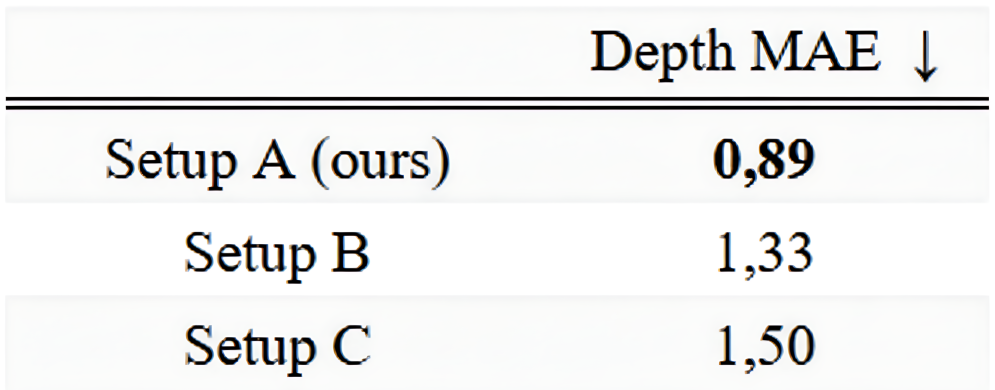} 
    \caption{Depth MAE $\downarrow$}
    \label{tab:abl_video}
\end{subfigure}
\hfill
\begin{subfigure}{0.48\columnwidth}
    \centering
    \includegraphics[width=\linewidth]{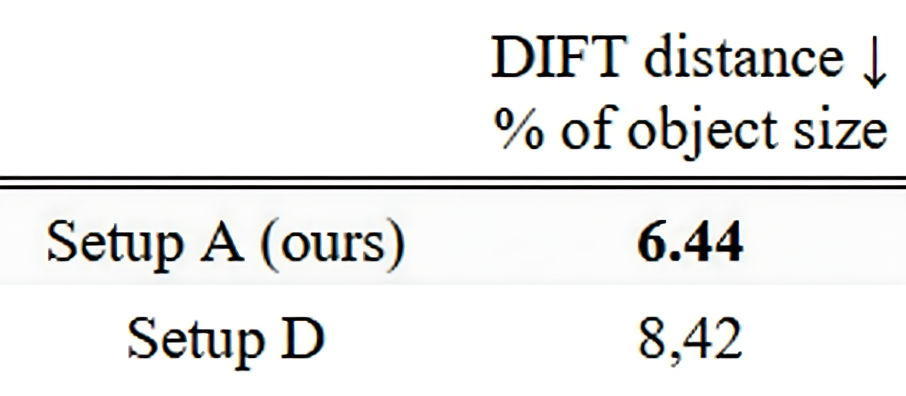}
    \caption{DIFT distance $\downarrow$ \\ \footnotesize (\% of object size)}
    \label{tab:abl_image}
\end{subfigure}

\caption{Ablation study results. Lower is better ($\downarrow$).}
\label{fig:ablation_vertical}
\end{table}
We evaluate setups (B) and (C) with the video pipeline reporting depth MAE structural score in Table~\ref{tab:abl_video}.
Since the setup (D) is roughly equivalent to MatchDiffusion, for which we have already reported results in Section~\ref{sec:experiments}, we decide to evaluate this setup with the image modality; the results are shown in Table~\ref{tab:abl_image}.
Visual examples for all the setups with the image pipeline are demonstrated in Figure~\ref{fig:ablation_study}.
Quantitative and qualitative comparisons demonstrate a gradual degradation of the results during the component removal.


\section{Conclusions}
We present a new universal method for joint inference with rectified flow models, which enables the rapid production of structurally aligned samples across different modalities.
Our method requires only a compact and local modification of the inference loop and can be applied on top of any pre-trained rectified flow model working with structured latent representations.
We demonstrate the performance of our method across three modalities—video, 3D, and images — comparing it both with editing-based methods and joint training methods.
Across all modalities, our method either achieves performance on par with the state of the art or surpasses it.
When applied to video modality, our method shows superior performance, enabling the accurate alignment of complex and vibrant environments. 
Our method has multiple potential applications, including the creation of aligned virtual environments and assets, and the generation of synthetic data.
In future work, the method can be further applied to other modalities, such as 4D video and keypoint movements.
\label{sec:conclusions}

\subsubsection*{Acknowledgments}
The work was supported by the grant for research centers in the field of AI provided by the Ministry of Economic Development of the Russian Federation in accordance with the agreement 000000C313925P4F0002 and the agreement №139-10-2025-033.

{
    \small
    \bibliographystyle{ieeenat_fullname}
    \bibliography{main}

@String(CVPR= {IEEE Conf. Comput. Vis. Pattern Recog.})

@String(ICCV= {Int. Conf. Comput. Vis.})

@String(TOG= {ACM Trans. Graph.})

@String(ICLR = {Int. Conf. Learn. Represent.})

@String(AAAI = {AAAI})

@String(CVPR  = {CVPR})

@String(ICCV  = {ICCV})

@String(TOG   = {ACM TOG})

@String(ICLR  = {ICLR})

@inproceedings{liu2023rectifiedflow,
  title        = {Flow Straight and Fast: Learning to Generate and Transfer Data with Rectified Flow},
  author       = {Liu, Xingchao and Gong, Chengyue and Liu, Qiang},
  booktitle    = {International Conference on Learning Representations (ICLR)},
  year         = {2023},
  url          = {https://openreview.net/forum?id=XVjTT1nw5z},
  eprint       = {2209.03003},
  archivePrefix= {arXiv},
  primaryClass = {cs.LG}
}

@article{ho2020ddpm,
  title={Denoising diffusion probabilistic models},
  author={Ho, Jonathan and Jain, Ajay and Abbeel, Pieter},
  journal={Advances in neural information processing systems},
  volume={33},
  pages={6840--6851},
  year={2020}
}

@article{song2020ddim,
  title={Denoising diffusion implicit models},
  author={Song, Jiaming and Meng, Chenlin and Ermon, Stefano},
  journal={arXiv preprint arXiv:2010.02502},
  year={2020}
}

@article{goodfellow2020gan,
  title={Generative adversarial networks},
  author={Goodfellow, Ian and Pouget-Abadie, Jean and Mirza, Mehdi and Xu, Bing and Warde-Farley, David and Ozair, Sherjil and Courville, Aaron and Bengio, Yoshua},
  journal={Communications of the ACM},
  volume={63},
  number={11},
  pages={139--144},
  year={2020},
  publisher={ACM New York, NY, USA}
}

@inproceedings{kang2023scalinggan,
  title={Scaling up gans for text-to-image synthesis},
  author={Kang, Minguk and Zhu, Jun-Yan and Zhang, Richard and Park, Jaesik and Shechtman, Eli and Paris, Sylvain and Park, Taesung},
  booktitle={Proceedings of the IEEE/CVF conference on computer vision and pattern recognition},
  pages={10124--10134},
  year={2023}
}

@article{huang2024deadgan,
  title={The gan is dead; long live the gan! a modern gan baseline},
  author={Huang, Nick and Gokaslan, Aaron and Kuleshov, Volodymyr and Tompkin, James},
  journal={Advances in Neural Information Processing Systems},
  volume={37},
  pages={44177--44215},
  year={2024}
}

@article{kingma2013vae,
  title={Auto-encoding variational bayes},
  author={Kingma, Diederik P and Welling, Max},
  journal={arXiv preprint arXiv:1312.6114},
  year={2013}
}

@article{van2017vqvae,
  title={Neural discrete representation learning},
  author={Van Den Oord, Aaron and Vinyals, Oriol and others},
  journal={Advances in neural information processing systems},
  volume={30},
  year={2017}
}

@inproceedings{rombach2022sd,
  title={High-resolution image synthesis with latent diffusion models},
  author={Rombach, Robin and Blattmann, Andreas and Lorenz, Dominik and Esser, Patrick and Ommer, Bj{\"o}rn},
  booktitle={Proceedings of the IEEE/CVF conference on computer vision and pattern recognition},
  pages={10684--10695},
  year={2022}
}

@article{ramesh2022dalle2,
  title={Hierarchical text-conditional image generation with clip latents},
  author={Ramesh, Aditya and Dhariwal, Prafulla and Nichol, Alex and Chu, Casey and Chen, Mark},
  journal={arXiv preprint arXiv:2204.06125},
  volume={1},
  number={2},
  pages={3},
  year={2022}
}

@article{saharia2022imagen,
  title={Photorealistic text-to-image diffusion models with deep language understanding},
  author={Saharia, Chitwan and Chan, William and Saxena, Saurabh and Li, Lala and Whang, Jay and Denton, Emily L and Ghasemipour, Kamyar and Gontijo Lopes, Raphael and Karagol Ayan, Burcu and Salimans, Tim and others},
  journal={Advances in neural information processing systems},
  volume={35},
  pages={36479--36494},
  year={2022}
}

@inproceedings{peebles2023dit,
  title={Scalable diffusion models with transformers},
  author={Peebles, William and Xie, Saining},
  booktitle={Proceedings of the IEEE/CVF international conference on computer vision},
  pages={4195--4205},
  year={2023}
}

@misc{blackforestlabs_flux1dev_2024,
  title        = {{FLUX}.1 [dev]},
  author       = {{Black Forest Labs}},
  year         = {2024},
  howpublished = {\url{https://huggingface.co/black-forest-labs/FLUX.1-dev}},
  note         = {Open-weight rectified-flow text-to-image model},
}

@inproceedings{esser2024sd3,
  title={Scaling rectified flow transformers for high-resolution image synthesis},
  author={Esser, Patrick and Kulal, Sumith and Blattmann, Andreas and Entezari, Rahim and M{\"u}ller, Jonas and Saini, Harry and Levi, Yam and Lorenz, Dominik and Sauer, Axel and Boesel, Frederic and others},
  booktitle={Forty-first international conference on machine learning},
  year={2024}
}

@misc{wu2025qwenimagetechnicalreport,
  title         = {Qwen-Image Technical Report},
  author        = {Chenfei Wu and Jiahao Li and Jingren Zhou and Junyang Lin and Kaiyuan Gao
                  and Kun Yan and Sheng-ming Yin and Shuai Bai and Xiao Xu and Yilei Chen
                  and Yuxiang Chen and Zecheng Tang and Zekai Zhang and Zhengyi Wang and
                  An Yang and Bowen Yu and Chen Cheng and Dayiheng Liu and Deqing Li and
                  Hang Zhang and Hao Meng and Wei Hu and Jingyuan Ni and Kai Chen and
                  Kuan Cao and Liang Peng and Lin Qu and Minggang Wu and Peng Wang and
                  Shuting Yu and Tingkun Wen and Wensen Feng and Xiaoxiao Xu and Yi Wang
                  and Yichang Zhang and Yongqiang Zhu and Yujia Wu and Yuxuan Cai and
                  Zenan Liu},
  year          = {2025},
  eprint        = {2508.02324},
  archivePrefix = {arXiv},
  primaryClass  = {cs.CV},
  url           = {https://arxiv.org/abs/2508.02324},
}

@inproceedings{rout2025semanticedit,
  title     = {Semantic Image Inversion and Editing using Rectified Stochastic Differential Equations},
  author    = {Rout, Litu and Chen, Yujia and Ruiz, Nataniel and Caramanis, Constantine and Shakkottai, Sanjay and Chu, Wen-Sheng},
  booktitle = {Proceedings of the Thirteenth International Conference on Learning Representations},
  year      = {2025},
  url       = {https://openreview.net/forum?id=Hu0FSOSEyS}
}

@inproceedings{Benarous2025SPIE,
  title={SPIE: Semantic and Structural Post-Training of Image Editing Diffusion Models with AI feedback},
  author={Benarous, Elior and Du, Yilun and Yang, Heng},
  booktitle={Proceedings of the IEEE/CVF International Conference on Computer Vision},
  pages={6395--6407},
  year={2025}
}

@misc{qwen_image_edit_2025,
  title        = {Qwen-Image-Edit},
  author       = {{Qwen Team}},
  year         = {2025},
  howpublished = {\url{https://huggingface.co/Qwen/Qwen-Image-Edit}},
  note         = {Image editing foundation model based on Qwen-Image},
}

@article{labs2025fluxkontext,
  title={FLUX. 1 Kontext: Flow Matching for In-Context Image Generation and Editing in Latent Space},
  author={Labs, Black Forest and Batifol, Stephen and Blattmann, Andreas and Boesel, Frederic and Consul, Saksham and Diagne, Cyril and Dockhorn, Tim and English, Jack and English, Zion and Esser, Patrick and others},
  journal={arXiv preprint arXiv:2506.15742},
  year={2025}
}

@inproceedings{vace,
    title     = {VACE: All-in-One Video Creation and Editing},
    author    = {Jiang, Zeyinzi and Han, Zhen and Mao, Chaojie and Zhang, Jingfeng and Pan, Yulin and Liu, Yu},
    booktitle = {Proceedings of the IEEE/CVF International Conference on Computer Vision},
    pages     = {17191-17202},
    year      = {2025}
}

@article{wan2025open,
  title   = {Wan: Open and Advanced Large-Scale Video Generative Models},
  author  = {{Team Wan} and Wang, Ang and Ai, Baole and Wen, Bin and Mao, Chaojie
             and Xie, Chen-Wei and Chen, Di and Yu, Feiwu and Zhao, Haiming
             and Yang, Jianxiao and others},
  journal = {arXiv preprint arXiv:2503.20314},
  year    = {2025}
}

@misc{decart2025lucyedit,
  title   = {Lucy Edit: Open-Weight Text-Guided Video Editing},
  author  = {{DecartAI Team}},
  year    = {2025},
  note    = {Technical report},
  url     = {https://d2drjpuinn46lb.cloudfront.net/Lucy_Edit__High_Fidelity_Text_Guided_Video_Editing.pdf}
}

@inproceedings{Pardo2025MatchDiffusion,
  title     = {MatchDiffusion: Training-free Generation of Match-Cuts},
  author    = {Pardo, Alejandro and Pizzati, Fabio and Zhang, Tong and Pondaven, Alexander and Torr, Philip and Perez, Juan Camilo and Ghanem, Bernard},
  booktitle = {Proceedings of the IEEE/CVF International Conference on Computer Vision (ICCV)},
  year      = {2025},
  month     = {October}
}

@article{shi2023mvdream,
  title={Mvdream: Multi-view diffusion for 3d generation},
  author={Shi, Yichun and Wang, Peng and Ye, Jianglong and Long, Mai and Li, Kejie and Yang, Xiao},
  journal={arXiv preprint arXiv:2308.16512},
  year={2023}
}

@article{li2023instant3d,
  title={Instant3d: Fast text-to-3d with sparse-view generation and large reconstruction model},
  author={Li, Jiahao and Tan, Hao and Zhang, Kai and Xu, Zexiang and Luan, Fujun and Xu, Yinghao and Hong, Yicong and Sunkavalli, Kalyan and Shakhnarovich, Greg and Bi, Sai},
  journal={arXiv preprint arXiv:2311.06214},
  year={2023}
}

@inproceedings{go2025splatflow,
  title={Splatflow: Multi-view rectified flow model for 3d gaussian splatting synthesis},
  author={Go, Hyojun and Park, Byeongjun and Jang, Jiho and Kim, Jin-Young and Kwon, Soonwoo and Kim, Changick},
  booktitle={Proceedings of the Computer Vision and Pattern Recognition Conference},
  pages={21524--21536},
  year={2025}
}

@inproceedings{huang2025mvadapter,
  title={Mv-adapter: Multi-view consistent image generation made easy},
  author={Huang, Zehuan and Guo, Yuan-Chen and Wang, Haoran and Yi, Ran and Ma, Lizhuang and Cao, Yan-Pei and Sheng, Lu},
  booktitle={Proceedings of the IEEE/CVF International Conference on Computer Vision},
  pages={16377--16387},
  year={2025}
}

@inproceedings{szymanowicz2025bolt3d,
  title={Bolt3d: Generating 3d scenes in seconds},
  author={Szymanowicz, Stanislaw and Zhang, Jason Y and Srinivasan, Pratul and Gao, Ruiqi and Brussee, Arthur and Holynski, Aleksander and Martin-Brualla, Ricardo and Barron, Jonathan T and Henzler, Philipp},
  booktitle={Proceedings of the IEEE/CVF International Conference on Computer Vision},
  pages={24846--24857},
  year={2025}
}

@article{poole2022dreamfusion,
  title={Dreamfusion: Text-to-3d using 2d diffusion},
  author={Poole, Ben and Jain, Ajay and Barron, Jonathan T and Mildenhall, Ben},
  journal={arXiv preprint arXiv:2209.14988},
  year={2022}
}

@article{liang2023luciddreamerism,
  title   = {LucidDreamer: Towards High-Fidelity Text-to-3D Generation via Interval Score Matching},
  author  = {Liang, Yixun and Yang, Xin and Lin, Jiantao and Li, Haodong
             and Xu, Xiaogang and Chen, Yingcong},
  journal = {arXiv preprint arXiv:2311.11284},
  year    = {2023}
}

@article{lukoianov2024sdi,
  title={Score distillation via reparametrized ddim},
  author={Lukoianov, Artem and S{\'a}ez de Oc{\'a}riz Borde, Haitz and Greenewald, Kristjan and Guizilini, Vitor and Bagautdinov, Timur and Sitzmann, Vincent and Solomon, Justin M},
  journal={Advances in Neural Information Processing Systems},
  volume={37},
  pages={26011--26044},
  year={2024}
}

@article{xiang2024trellis,
  title   = {Structured 3D Latents for Scalable and Versatile 3D Generation},
  author  = {Xiang, Jianfeng and Lv, Zelong and Xu, Sicheng and Deng, Yu and Wang, Ruicheng and Zhang, Bowen and Chen, Dong and Tong, Xin and Yang, Jiaolong},
  journal = {arXiv preprint arXiv:2412.01506},
  year    = {2024}
}

@article{xiang2025native,
  title={Native and compact structured latents for 3d generation},
  author={Xiang, Jianfeng and Chen, Xiaoxue and Xu, Sicheng and Wang, Ruicheng and Lv, Zelong and Deng, Yu and Zhu, Hongyuan and Dong, Yue and Zhao, Hao and Yuan, Nicholas Jing and others},
  journal={arXiv preprint arXiv:2512.14692},
  year={2025}
}

@article{wu2025unilat3d,
  title={UniLat3D: Geometry-Appearance Unified Latents for Single-Stage 3D Generation},
  author={Wu, Guanjun and Fang, Jiemin and Yang, Chen and Li, Sikuang and Yi, Taoran and Lu, Jia and Zhou, Zanwei and Cen, Jiazhong and Xie, Lingxi and Zhang, Xiaopeng and others},
  journal={arXiv preprint arXiv:2509.25079},
  year={2025}
}

@inproceedings{yang2025prometheus,
  title={Prometheus: 3d-aware latent diffusion models for feed-forward text-to-3d scene generation},
  author={Yang, Yuanbo and Shao, Jiahao and Li, Xinyang and Shen, Yujun and Geiger, Andreas and Liao, Yiyi},
  booktitle={Proceedings of the Computer Vision and Pattern Recognition Conference},
  pages={2857--2869},
  year={2025}
}

@article{zhao2025hunyuan3d,
  title={Hunyuan3d 2.0: Scaling diffusion models for high resolution textured 3d assets generation},
  author={Zhao, Zibo and Lai, Zeqiang and Lin, Qingxiang and Zhao, Yunfei and Liu, Haolin and Yang, Shuhui and Feng, Yifei and Yang, Mingxin and Zhang, Sheng and Yang, Xianghui and others},
  journal={arXiv preprint arXiv:2501.12202},
  year={2025}
}

@article{lai2025hunyuan3d25,
  title={Hunyuan3D 2.5: Towards High-Fidelity 3D Assets Generation with Ultimate Details},
  author={Lai, Zeqiang and Zhao, Yunfei and Liu, Haolin and Zhao, Zibo and Lin, Qingxiang and Shi, Huiwen and Yang, Xianghui and Yang, Mingxin and Yang, Shuhui and Feng, Yifei and others},
  journal={arXiv preprint arXiv:2506.16504},
  year={2025}
}

@article{lai2025unleashing,
  title={Unleashing vecset diffusion model for fast shape generation},
  author={Lai, Zeqiang and Zhao, Yunfei and Zhao, Zibo and Liu, Haolin and Wang, Fuyun and Shi, Huiwen and Yang, Xianghui and Lin, Qingxiang and Huang, Jingwei and Liu, Yuhong and others},
  journal={arXiv preprint arXiv:2503.16302},
  year={2025}
}

@article{zhuang2024tip,
  title={Tip-editor: An accurate 3d editor following both text-prompts and image-prompts},
  author={Zhuang, Jingyu and Kang, Di and Cao, Yan-Pei and Li, Guanbin and Lin, Liang and Shan, Ying},
  journal={ACM Transactions on Graphics (TOG)},
  volume={43},
  number={4},
  pages={1--12},
  year={2024},
  publisher={ACM New York, NY, USA}
}

@inproceedings{li2024focaldreamer,
  title={Focaldreamer: Text-driven 3d editing via focal-fusion assembly},
  author={Li, Yuhan and Dou, Yishun and Shi, Yue and Lei, Yu and Chen, Xuanhong and Zhang, Yi and Zhou, Peng and Ni, Bingbing},
  booktitle={Proceedings of the AAAI conference on artificial intelligence},
  volume={38},
  number={4},
  pages={3279--3287},
  year={2024}
}

@inproceedings{zhuang2023dreameditor,
  title={Dreameditor: Text-driven 3d scene editing with neural fields},
  author={Zhuang, Jingyu and Wang, Chen and Lin, Liang and Liu, Lingjie and Li, Guanbin},
  booktitle={SIGGRAPH Asia 2023 Conference Papers},
  pages={1--10},
  year={2023}
}

@article{chung2023luciddreamer,
  title={LucidDreamer: Domain-free Generation of 3D Gaussian Splatting Scenes},
  author={Chung, Jaeyoung and Lee, Suyoung and Nam, Hyeongjin and Lee, Jaerin and Lee, Kyoung Mu},
  journal={arXiv preprint arXiv:2311.13384},
  year={2023}
}

@article{mvedit2024,
  title   = {Generic 3D Diffusion Adapter Using Controlled Multi-View Editing},
  author  = {Chen, Hansheng and Shi, Ruoxi and Liu, Yulin and Shen, Bokui
             and Gu, Jiayuan and Wetzstein, Gordon and Su, Hao and Guibas, Leonidas J.},
  journal = {arXiv preprint arXiv:2403.12032},
  year    = {2024}
}

@inproceedings{chen2024dge,
  title={Dge: Direct gaussian 3d editing by consistent multi-view editing},
  author={Chen, Minghao and Laina, Iro and Vedaldi, Andrea},
  booktitle={European Conference on Computer Vision},
  pages={74--92},
  year={2024},
  organization={Springer}
}

@inproceedings{lee2025editsplat,
  title={Editsplat: Multi-view fusion and attention-guided optimization for view-consistent 3d scene editing with 3d gaussian splatting},
  author={Lee, Dong In and Park, Hyeongcheol and Seo, Jiyoung and Park, Eunbyung and Park, Hyunje and Baek, Ha Dam and Shin, Sangheon and Kim, Sangmin and Kim, Sangpil},
  booktitle={Proceedings of the Computer Vision and Pattern Recognition Conference},
  pages={11135--11145},
  year={2025}
}

@article{li2025voxhammer,
    title = {VoxHammer: Training-Free Precise and Coherent 3D Editing in Native 3D Space},
    author = {Li, Lin and Huang, Zehuan and Feng, Haoran and Zhuang, Gengxiong and Chen, Rui and Guo, Chunchao and Sheng, Lu},
    journal = {arXiv preprint arXiv:2508.19247},
    year = {2025}
}

@article{ye2025nano3d,
  title={NANO3D: A Training-Free Approach for Efficient 3D Editing Without Masks},
  author={Ye, Junliang and Xie, Shenghao and Zhao, Ruowen and Wang, Zhengyi and Yan, Hongyu and Zu, Wenqiang and Ma, Lei and Zhu, Jun},
  journal={arXiv preprint arXiv:2510.15019},
  year={2025}
}

@article{lee2023syncdiffusion,
  title={Syncdiffusion: Coherent montage via synchronized joint diffusions},
  author={Lee, Yuseung and Kim, Kunho and Kim, Hyunjin and Sung, Minhyuk},
  journal={Advances in Neural Information Processing Systems},
  volume={36},
  pages={50648--50660},
  year={2023}
}

@article{yeo2025stochsync,
  title={StochSync: Stochastic Diffusion Synchronization for Image Generation in Arbitrary Spaces},
  author={Yeo, Kyeongmin and Kim, Jaihoon and Sung, Minhyuk},
  journal={arXiv preprint arXiv:2501.15445},
  year={2025}
}

@inproceedings{Ignatyev2025A3D,
  title     = {A3D: Does Diffusion Dream about {3D} Alignment?},
  author    = {Ignatyev, Savva Victorovich and Konovalova, Nina and Selikhanovych, Daniil and Voynov, Oleg and Patakin, Nikolay and Olkov, Ilya and Senushkin, Dmitry and Artemov, Alexey and Konushin, Anton and Filippov, Alexander and Wonka, Peter and Burnaev, Evgeny},
  booktitle = {International Conference on Learning Representations (ICLR)},
  year      = {2025},
  url       = {https://openreview.net/forum?id=QQCIfkhGIq},
  note      = {Poster}
}

@article{depth_anything_v2,
  title={Depth Anything V2},
  author={Yang, Lihe and Kang, Bingyi and Huang, Zilong and Zhao, Zhen and Xu, Xiaogang and Feng, Jiashi and Zhao, Hengshuang},
  journal={arXiv:2406.09414},
  year={2024}
}

@article{ren2024grounded,
  title   = {Grounded SAM: Assembling Open-World Models for Diverse Visual Tasks},
  author  = {Ren, Tianhe and Liu, Shilong and Zeng, Ailing and Lin, Jing and Li, Kunchang and Cao, He and Chen, Jiayu and Huang, Xinyu and Chen, Yukang and Yan, Feng and Zeng, Zhaoyang and Zhang, Hao and Li, Feng and Yang, Jie and Li, Hongyang and Jiang, Qing and Zhang, Lei},
  journal = {arXiv preprint arXiv:2401.14159},
  year    = {2024}
}

@inproceedings{tang2023emergent,
  title        = {Emergent Correspondence from Image Diffusion},
  author       = {Tang, Luming and Jia, Menglin and Wang, Qianqian and Phoo, Cheng Perng and Hariharan, Bharath},
  booktitle    = {Advances in Neural Information Processing Systems 36 (NeurIPS 2023)},
  year         = {2023},
  url          = {https://openreview.net/forum?id=ypOiXjdfnU}
}

@inproceedings{
berthelot2018understanding,
title={Understanding and Improving Interpolation in Autoencoders via an Adversarial Regularizer},
author={David Berthelot* and Colin Raffel* and Aurko Roy and Ian Goodfellow},
booktitle={International Conference on Learning Representations},
year={2019},
url={https://openreview.net/forum?id=S1fQSiCcYm},
}

@INPROCEEDINGS{karras2020analyzing,
  author={Karras, Tero and Laine, Samuli and Aittala, Miika and Hellsten, Janne and Lehtinen, Jaakko and Aila, Timo},
  booktitle={2020 IEEE/CVF Conference on Computer Vision and Pattern Recognition (CVPR)}, 
  title={Analyzing and Improving the Image Quality of StyleGAN}, 
  year={2020},
  volume={},
  number={},
  pages={8107-8116},
  doi={10.1109/CVPR42600.2020.00813}}

@article{avrahami2023chosen,
  title   = {The Chosen One: Consistent Characters in Text-to-Image Diffusion Models},
  author  = {Avrahami, Omri and Hertz, Amir and Vinker, Yael and Arar, Moab and Fruchter, Shlomi and Fried, Ohad and Cohen-Or, Daniel and Lischinski, Dani},
  journal = {arXiv preprint arXiv:2311.10093},
  year    = {2023}
}

@InProceedings{Hertz_2024_CVPR,
  author    = {Hertz, Amir and Voynov, Andrey and Fruchter, Shlomi and Cohen-Or, Daniel},
  title     = {Style Aligned Image Generation via Shared Attention},
  booktitle = {Proceedings of the IEEE/CVF Conference on Computer Vision and Pattern Recognition (CVPR)},
  month     = {June},
  year      = {2024},
  pages     = {4775-4785}
}

@misc{pardo2024matchdiffusiontrainingfreegenerationmatchcuts,
      title={MatchDiffusion: Training-free Generation of Match-cuts}, 
      author={Alejandro Pardo and Fabio Pizzati and Tong Zhang and Alexander Pondaven and Philip Torr and Juan Camilo Perez and Bernard Ghanem},
      year={2024},
      eprint={2411.18677},
      archivePrefix={arXiv},
      primaryClass={cs.CV},
      url={https://arxiv.org/abs/2411.18677}, 
}

@misc{jiang2025vaceallinonevideocreation,
      title={VACE: All-in-One Video Creation and Editing}, 
      author={Zeyinzi Jiang and Zhen Han and Chaojie Mao and Jingfeng Zhang and Yulin Pan and Yu Liu},
      year={2025},
      eprint={2503.07598},
      archivePrefix={arXiv},
      primaryClass={cs.CV},
      url={https://arxiv.org/abs/2503.07598}, 
}

@misc{wan2025wanopenadvancedlargescale,
      title={Wan: Open and Advanced Large-Scale Video Generative Models}, 
      author={Team Wan and Ang Wang and Baole Ai and Bin Wen and Chaojie Mao and Chen-Wei Xie and Di Chen and Feiwu Yu and Haiming Zhao and Jianxiao Yang and Jianyuan Zeng and Jiayu Wang and Jingfeng Zhang and Jingren Zhou and Jinkai Wang and Jixuan Chen and Kai Zhu and Kang Zhao and Keyu Yan and Lianghua Huang and Mengyang Feng and Ningyi Zhang and Pandeng Li and Pingyu Wu and Ruihang Chu and Ruili Feng and Shiwei Zhang and Siyang Sun and Tao Fang and Tianxing Wang and Tianyi Gui and Tingyu Weng and Tong Shen and Wei Lin and Wei Wang and Wei Wang and Wenmeng Zhou and Wente Wang and Wenting Shen and Wenyuan Yu and Xianzhong Shi and Xiaoming Huang and Xin Xu and Yan Kou and Yangyu Lv and Yifei Li and Yijing Liu and Yiming Wang and Yingya Zhang and Yitong Huang and Yong Li and You Wu and Yu Liu and Yulin Pan and Yun Zheng and Yuntao Hong and Yupeng Shi and Yutong Feng and Zeyinzi Jiang and Zhen Han and Zhi-Fan Wu and Ziyu Liu},
      year={2025},
      eprint={2503.20314},
      archivePrefix={arXiv},
      primaryClass={cs.CV},
      url={https://arxiv.org/abs/2503.20314}, 
}

@misc{hong2022cogvideolargescalepretrainingtexttovideo,
      title={CogVideo: Large-scale Pretraining for Text-to-Video Generation via Transformers}, 
      author={Wenyi Hong and Ming Ding and Wendi Zheng and Xinghan Liu and Jie Tang},
      year={2022},
      eprint={2205.15868},
      archivePrefix={arXiv},
      primaryClass={cs.CV},
      url={https://arxiv.org/abs/2205.15868}, 
}

@misc{ju2025editverseunifyingimagevideo,
      title={EditVerse: Unifying Image and Video Editing and Generation with In-Context Learning}, 
      author={Xuan Ju and Tianyu Wang and Yuqian Zhou and He Zhang and Qing Liu and Nanxuan Zhao and Zhifei Zhang and Yijun Li and Yuanhao Cai and Shaoteng Liu and Daniil Pakhomov and Zhe Lin and Soo Ye Kim and Qiang Xu},
      year={2025},
      eprint={2509.20360},
      archivePrefix={arXiv},
      primaryClass={cs.CV},
      url={https://arxiv.org/abs/2509.20360}, 
}

@misc{nanobanana,
  title        = {Image editing in Gemini just got a major upgrade},
  author       = {{Google}},
  year         = {2025},
  month        = aug,
  url          = {https://blog.google/products/gemini/updated-image-editing-model/},
  note         = {Describes \emph{Nano Banana} (Gemini 2.5 Flash Image). Accessed: 2025-11-13}
}

@inproceedings{Wu2024GPTEval3D,
  author    = {Tong Wu and Guandao Yang and Zhibing Li and Kai Zhang and Ziwei Liu and Leonidas J. Guibas and Dahua Lin and Gordon Wetzstein},
  title     = {{GPT-4V(ision) is a Human-Aligned Evaluator for Text-to-3D Generation}},
  booktitle = {Proceedings of the IEEE/CVF Conference on Computer Vision and Pattern Recognition (CVPR)},
  year      = {2024},
  url       = {https://openaccess.thecvf.com/content/CVPR2024/html/Wu_GPT-4Vision_is_a_Human-Aligned_Evaluator_for_Text-to-3D_Generation_CVPR_2024_paper.html}
}

@article{huang2025t2icompbench++,
  author  = {Huang, Kaiyi and Duan, Chengqi and Sun, Kaiyue and Xie, Enze and Li, Zhenguo and Liu, Xihui},
  title   = {T2I-CompBench++: An Enhanced and Comprehensive Benchmark for Compositional Text-to-Image Generation},
  journal = {IEEE Transactions on Pattern Analysis and Machine Intelligence},
  year    = {2025},
  pages   = {1--17},
  url     = {https://doi.ieeecomputersociety.org/10.1109/TPAMI.2025.3531907},
  publisher = {IEEE Computer Society},
  address   = {Los Alamitos, CA, USA},
  month   = jan,
  issn    = {1939-3539}
}
}
\maketitlesupplementary



\definecolor{cvprblue}{rgb}{0.21,0.49,0.74}



\def\paperID{*****} 
\def\confName{CVPR}
\def\confYear{2026}



\maketitle
\appendix



\section{Competitor details}

\subsection{2D competitors}

For running competitors, we use the same pair of text prompts. Source prompt corresponds to the first prompt in the pair, and the target prompt corresponds to the second prompt in the pair.

\begin{enumerate}
    \item \textbf{Qwen-Image-Edit}. We use the most recent release of the model from 25.09. 
    The source image is generated by FLUX \cite{blackforestlabs_flux1dev_2024} using the source prompt. For target image generation, we modify the prompt from 'target prompt' to "change 'source prompt' to 'target prompt', preserving geometry and background" 
    \item \textbf{RF-inversion}. We generate the source image with the FLUX model \cite{blackforestlabs_flux1dev_2024}. Hyperparameters for target generation are $t_{stop}=6/28$, $\eta=0.9$, $n_{steps}=28$.
\end{enumerate}

\subsection{3D competitors}
For 3D evaluation, we use the 3D models, renders, and numerical results from A3D paper~\cite{Ignatyev2025A3D}.

\subsection{Video competitors}

\begin{enumerate}
    \item \textbf{LucyEdit}. We use a WAN model to generate source videos. We edit them using 2 keywords: 'replace' - for changing the main character on the scene, and 'transform' - for performing global scene-level edits.
    \item \textbf{VACE}. We use WAN to generate source videos. We edit them using Wan2.1-VACE-1.3B, conditioned on source video depth.
    \item \textbf{MatchDiffusion}. We use exactly the same prompts as for the main method, which are presented in the Table~\ref{sup:tab:video_promtps}.
\end{enumerate}

\section{Math details}

\subsection{Integral form}
In Section~4.1, we have derived the formulas for the single-step update for our method. 
While these formulas are analytically correct, the update depends on the placement and the number of the sampled points, and not only on the distribution $p(\alpha)$.
If we set $\lim_{k \to \infty}$ in Equation~3, we obtain the formulas in the integral form (omitting 
\begin{equation}
\label{sup:eq:loss}
    \mathcal{L}(x^a_{t_2}, x^b_{t_2})  = \int_{0}^{1} p(\alpha)\| x_{t_2}(\alpha) - \hat{x}_{t_2}(\alpha) \|^2 d \alpha,\\
\end{equation}
\[
    \hat{x}_{t_2}(\alpha) = x_{t_1}(\alpha) + (t_2 - t_1)\, v_{\Theta}\bigl(x_{t_1}(\alpha), t_1, c(\alpha)\bigr)\\
\]
\[
    x_{t_2}(\alpha) = (1-\alpha)\,x^a_{t_2} + \alpha\,x^b_{t_2}
\]
\[
    x_{t_1}(\alpha) = (1-\alpha)\,x^a_{t_1} + \alpha\,x^b_{t_1}
\]
We can also rewrite $\hat{x}_{t_2}(\alpha)$ in the differential form:
\begin{equation}
\label{eq:differential}
    \hat{x}_{t_2}(\alpha) = 
    x_{t_1}(\alpha) + 
    dt\, v_{\Theta}\bigl(x_{t_1}(\alpha), t_1, c(\alpha)\bigr)
\end{equation}

We can also rewrite the solution of the regression within the same logic. 
\[
c_{00} = \int_{\alpha=0}^1 p(\alpha) (1-\alpha)^2 d\alpha,\quad
c_{01} = \int_{\alpha=0}^1 p(\alpha)(1-\alpha)\alpha \, d\alpha,
\]
\[
c_{11} = \int_{\alpha=0}^1 p(\alpha)\alpha^2 \, d\alpha ,\quad
\Delta=c_{00}c_{11} - c_{01}^2,
\]
\[
d_0 = \int\limits_{\alpha=0}^1 p(\alpha)(1-\alpha)\hat x_{t_2}(\alpha) \, d\alpha ,\;\;
d_1 = \int\limits_{\alpha=0}^1 p(\alpha)\alpha \, \hat x_{t_2}(\alpha) \, d\alpha, \quad
\]

We can use Equation~\ref{eq:differential} to write out:

\begin{equation}
\begin{aligned}
d_0 = & \int_{\alpha=0}^1 p(\alpha)(1-\alpha)
x_{t_1}(\alpha)\, d\alpha \, + \\
& dt \int_{\alpha=0}^1 \, p(\alpha)(1-\alpha) \, v_{\Theta}\bigl(x_{t_1}(\alpha), t_1, c(\alpha)\bigr) \, d\alpha ,\quad \\
d_1 = &\int_{\alpha=0}^{1} p(\alpha) \, \alpha \, x_{t_1}(\alpha)  d\alpha + \\
& dt \int_{\alpha=0}^{1} p(\alpha) \, \alpha \, v_{\Theta}\bigl(x_{t_1}(\alpha), t_1, c(\alpha)\bigr)  d\alpha
\end{aligned}
\end{equation}


Here we split the integral into the sum of two integrals, where only the second integral is dependent on velocity and $dt$. 
Using
\begin{equation}
\label{eq:solution}
x_{t_2}^a = \frac{c_{11} d_0 - c_{01} d_1}{\Delta}, \quad
x_{t_2}^b = \frac{c_{00} d_1 - c_{01} d_0}{\Delta},
\end{equation}
After simplification, we derive:
\[
\begin{aligned}
x^a_{t_1+dt} = x^a_{t_1} + \frac{dt}{\Delta}\left( c_{11}\mu_0 - c_{01}\mu_1 \right) \\[6pt]
x^b_{t_1+dt} = x^b_{t_1} + \frac{dt}{\Delta}\left( c_{00}\mu_1 - c_{01}\mu_0 \right)
\end{aligned}
\]

\begin{equation}
\label{sup:eq:integral_solution}
\begin{aligned}
\mu_0 & = \int_{0}^{1} p(\alpha)(1-\alpha)\, v_{\Theta}\bigl(x_{t_1}(\alpha), t_1, c(\alpha)\bigr)\, d\alpha \\
\mu_1 & = \int_{0}^{1} p(\alpha)\alpha\, v_{\Theta}\bigl(x_{t_1}(\alpha), t_1, c(\alpha)\bigr)\, d\alpha
\end{aligned}
\end{equation}
Now we can write out $v_{t1}^a$ and $v_{t1}^b$. It can be seen that it only depends on the velocity-weighted closed integral.
\begin{equation}
    v^a_{t1} = \frac{c_{11}\mu_0 - c_{01}\mu_1}{\Delta} \quad
    v^b_{t1} = \frac{c_{00}\mu_1 - c_{01}\mu_0}{\Delta}
\end{equation}
We have derived the precise continuous version for the conversion of the rectified flow into the velocity field for segments. 
Note that we can use different approximations to calculate the integrals for $\mu_0$ and $\mu_1$, not necessarily using an even grid. 
For example, we can employ the Monte-Carlo method using the probability $p(\alpha)$ to sample points randomly from $\alpha \in [0, 1]$.
\subsection{Probability matching}
\label{sup:section:probability_matching}
We can show that under certain restrictions, our algorithm can be seen as conserving the probability distribution of the points moving along the $v_\theta$ trajectories.
Our primary goal is to conserve the probability distribution $p(\alpha)$. 

We define the target and approximate noisy marginals as
\begin{align}
p_t^\text{true}(x)
&= \int_0^1 p(\alpha)\,
   \mathcal{N}\bigl(x \mid x_t^\text{true}(\alpha), \sigma^2 I\bigr)\,d\alpha,\\
p_t^\text{approx}(x)
&= \int_0^1 p(\alpha)\,
   \mathcal{N}\bigl(x \mid x_t^\text{approx}(\alpha), \sigma^2 I\bigr)\,d\alpha.
\end{align}
It is convenient to consider the joint distributions over $(x,\alpha)$:
\begin{align}
p_t^\text{true}(x,\alpha)
&= p(\alpha)\,
   \mathcal{N}\bigl(x \mid x_t^\text{true}(\alpha), \sigma^2 I\bigr),\\
p_t^\text{approx}(x,\alpha)
&= p(\alpha)\,
   \mathcal{N}\bigl(x \mid x_t^\text{approx}(\alpha), \sigma^2 I\bigr).
\end{align}
For fixed $\alpha$, both conditionals $p_t^\text{true}(x\mid\alpha)$ and
$p_t^\text{approx}(x\mid\alpha)$ are Gaussians with the same covariance
$\sigma^2 I$, so their conditional KL divergence is
\begin{align}
\mathrm{KL}\bigl(
  p_t^\text{true}(x\mid\alpha)\,&\|\,p_t^\text{approx}(x\mid\alpha)
\bigr)
=\\&= \frac{1}{2\sigma^2}\,
  \bigl\|x_t^\text{true}(\alpha) - x_t^\text{approx}(\alpha)\bigr\|^2.
\end{align}
Taking the expectation with respect to $p(\alpha)$ yields the joint KL
\begin{align}
\mathrm{KL}\bigl(p_t^\text{true}(x,\alpha)\,&\|\,p_t^\text{approx}(x,\alpha)\bigr)
=\\ &= \frac{1}{2\sigma^2}\,
  \mathbb{E}_{\alpha\sim p}\bigl\|
    x_t^\text{true}(\alpha) - x_t^\text{approx}(\alpha)
  \bigr\|^2.\notag
\end{align}
As $\sigma \to 0$, the KL divergence between the marginals
$p_t^\text{true}(x)$ and $p_t^\text{approx}(x)$ differs from the joint KL
only by an $\mathcal{O}(1)$ term, so we obtain the asymptotic
\begin{align}
\label{sup:eq:KL}
\mathrm{KL}&\bigl(p_t^\text{true}\,\|\,p_t^\text{approx}\bigr)
=\\&= \frac{1}{2\sigma^2}\,
  \mathbb{E}_{\alpha\sim p}\bigl\|
    x_t^\text{true}(\alpha) - x_t^\text{approx}(\alpha)
  \bigr\|^2
+ \mathcal{O}(1),
\;\; \sigma \to 0.\notag
\end{align}
Thus, minimizing the $L^2$ objective
\begin{equation}
\mathbb{E}_{\alpha\sim p}\bigl\|
  x_t^\text{true}(\alpha) - x_t^\text{approx}(\alpha)
\bigr\|^2,
\end{equation}
Eq.~\ref{sup:eq:KL}, is equivalent to minimizing
$\mathrm{KL}\bigl(p_t^\text{true}\,\|\,p_t^\text{approx}\bigr)$ at leading
order in $\sigma^{-2}$.

\subsection{Plausibility optimization}
\label{subsec:likelihood_optimization}

Our segment-based rectified flow formulation can be interpreted as optimizing the log-likelihood of all samples along the segment, not just its endpoints. This perspective connects our approach to the fundamental likelihood maximization principle in continuous normalizing flows.

We have shown that for small $\sigma$, $p_{t_2}^{\text{true}}(x)$ concentrates around the true trajectory $\hat{x}_{t_2}(\alpha)$. 
Thus, minimizing $\mathcal{L}$ directly maximizes the expected log-likelihood of all points along our segment under the true data distribution.




Any sample drawn from a point along the segment (according to $p(\alpha)$) will have high likelihood under the target distribution, preserving the core property of rectified flows while maintaining the geometric structure of segments.

\section{User study details}\label{sec:ablation_user_study}
In the user study, annotators were asked to evaluate generated video pairs based on three criteria:

\begin{enumerate}[label=\textbf{\arabic*.}]
    \item \textbf{Alignment.} \textit{In which row are Video A and Video B better aligned with each other in terms of overall structure, overall meaning, pose, and 3D geometry?}
    
    \item \textbf{Visual Appeal.} \textit{In which row is the pair Video A and Video B more visually appealing in terms of realism, smoothness, and overall perceptual quality?}
    
    \item \textbf{Text Prompt Consistency.} \textit{In which row do Video A and Video B better match their textual description?}
\end{enumerate}

\begin{figure*}[h!]
    \includegraphics[width=\textwidth]{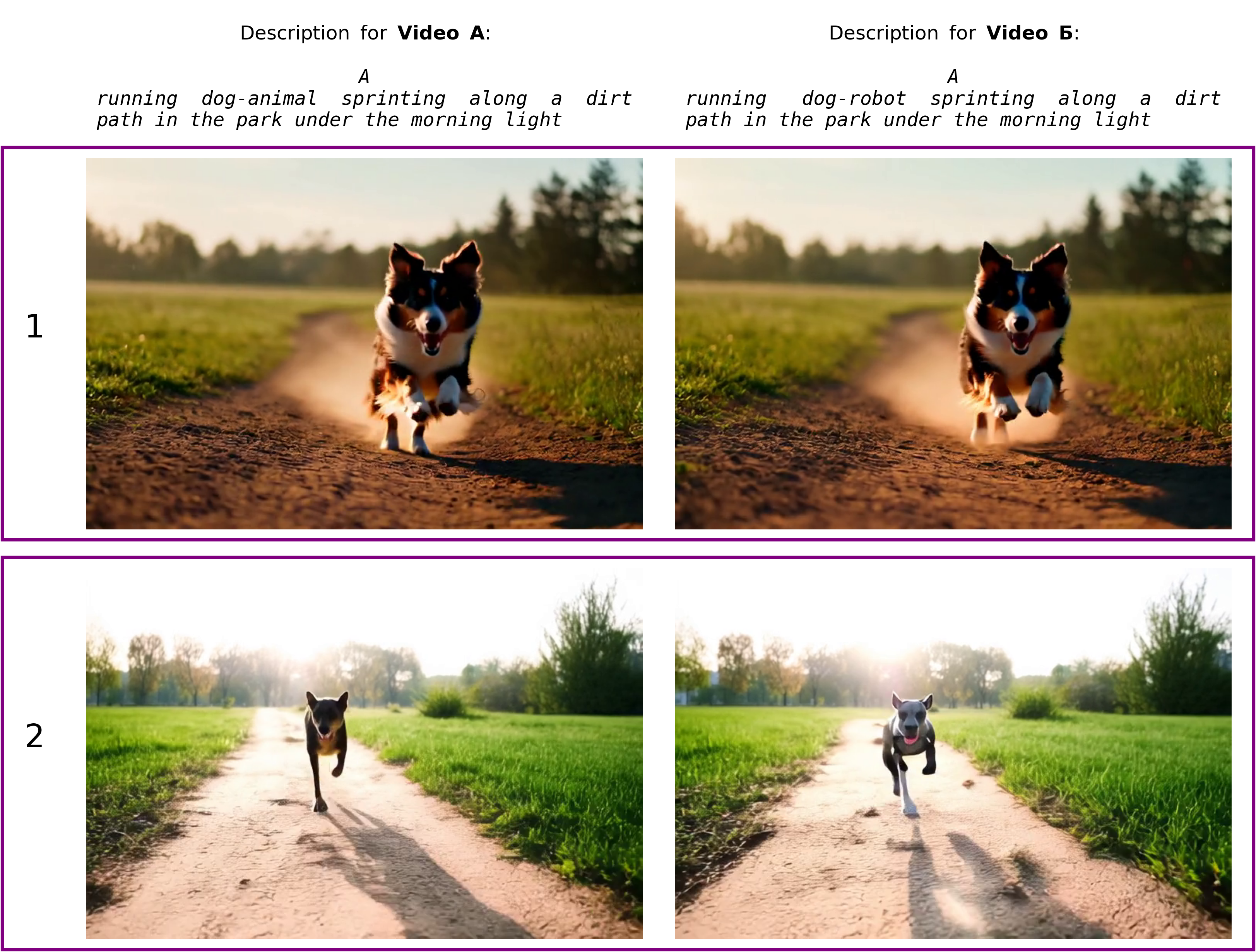}
    \caption{User study image example.}
    \label{fig:fig_us}
\end{figure*}

Figure~\ref{fig:fig_us} shows an example task from the study. All videos in the task were played simultaneously for the annotator with the ability to view them frame-by-frame.

For each question, annotators could choose between three options: preference for the first row, preference for the second row, or no preference.

We generated examples for the User Study using $21$ pairs from a diverse set of scenes described in Table~\ref{sup:tab:video_promtps}. To reduce bias, the rows were randomly swapped. At least seven different individuals annotated each task, and their responses were aggregated for analysis, resulting in a total of $ 2379$ answers from $60$ unique users across $63$ tasks.

\section{Full ablation}
\subsection{Method ablation}

We provide the detailed ablation experiments with image modality in Table~\ref{sup:tab:full_ablation}.
We also provide the version of the setup (D) without finding the best-performing hyperparameter (the number of joint steps), which is obtained directly by removing the component from the setup (C).
In this way, we show that the removal of the essential components leads to consistent degradation of the essential metrics. The only exception is the improvement observed when moving from setup (C) to setup (D). Sampling intermediate points $x_{t. i}$ and setting  $v_t^{anchor} = v_{\theta}(\frac{x_t^a + x_t^b}{2})$ are strongly interconnected and were both proposed together for joint segment transport. Consequently, using only intermediate points sampling alone can be less effective and even detrimental. Our primary aim here is not to optimize each intermediate variant, but to demonstrate that the endpoint of this sequential component removal—corresponding to the MatchDiffusion setup—exhibits degraded performance compared to our full method.

\begin{table}[t]
\vspace{-6pt}
\centering
\footnotesize
\begin{tabular}{lcccc}
\hline
\textbf{Number of sampled points} & 4 & 6 & 8 & 16 \\
\hline
Average DIFT score $\downarrow$ & 6.5 & 5.2 & \underline{4.8} & \textbf{3.2} \\
Average Depth L1 $\downarrow$   & 28.1 & 24.2 & \underline{22.2} & \textbf{16.2} \\
\hline
\end{tabular}
\vspace{-6pt}
\caption{\scriptsize Influence of the number of sampled points on performance.}
\vspace{-6pt}
\vspace{-6pt}
\label{tab:sampling_points}
\end{table}

\begin{table*}[t]
    \centering
    \caption{Full Ablation.}
    \label{sup:tab:full_ablation}
    \includegraphics[scale=0.33]{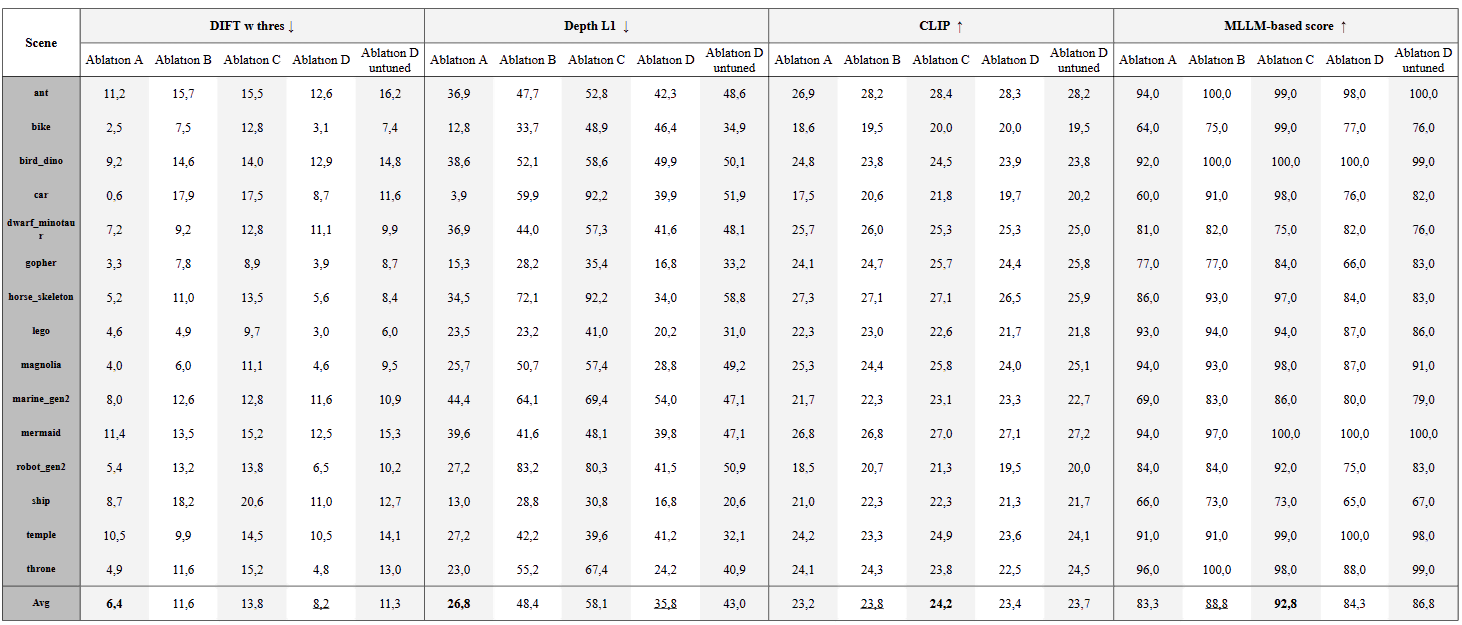}
\end{table*}

\subsection{Hyperparameter sensibility}

\begin{figure*}[h]
    \centering
    \includegraphics[width=0.95\linewidth]{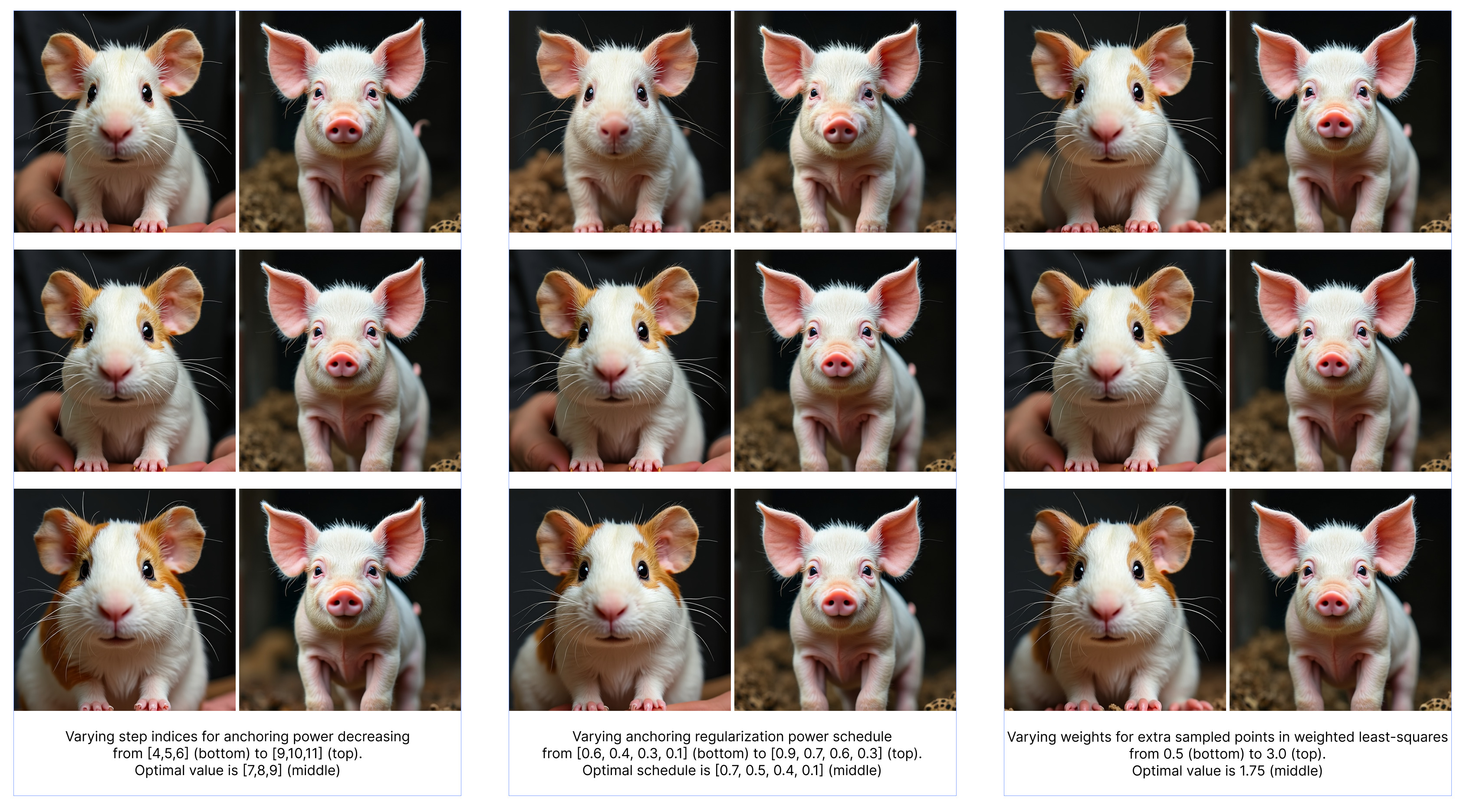}
    \caption{Visualization of hyperparameter sensitivity.}
    \label{fig:sens}
\end{figure*}

Our algorithm's hyperparameters such as anchoring regularization weight and probability distribution $p(\alpha)$ allow user-controlled trade-offs. Figure~\ref{fig:sens} shows that tuning the alignment parameters $w(t)$ and $p(\alpha)$ controls the balance between spatial alignment and text-image fidelity.

Another one hyperparameter is number of points sampled from $p(\alpha)$. This hyperparameter controls how precise the Monte-Carlo approximation of the probability integral along segment is. We demonstrate in Table \ref{tab:sampling_points} that increasing the number of sample points on the transition segment yields even higher semantic alignment.

\section{Additional image generation results}
\newcommand{\subfigwidth}{0.3\textwidth} 

\begin{figure*}[ht]
  \centering
  \begin{subfigure}[t]{\subfigwidth}
    \includegraphics[width=\textwidth]{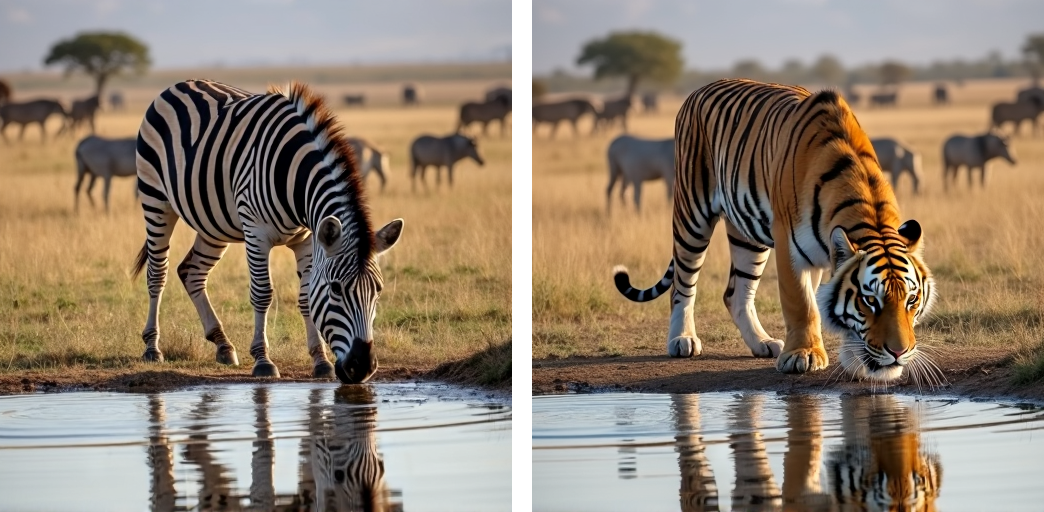}
    \caption{A zebra is drinking vs A tiger is drinking}
  \end{subfigure}%
  \hfill
  \begin{subfigure}[t]{\subfigwidth}
    \includegraphics[width=\textwidth]{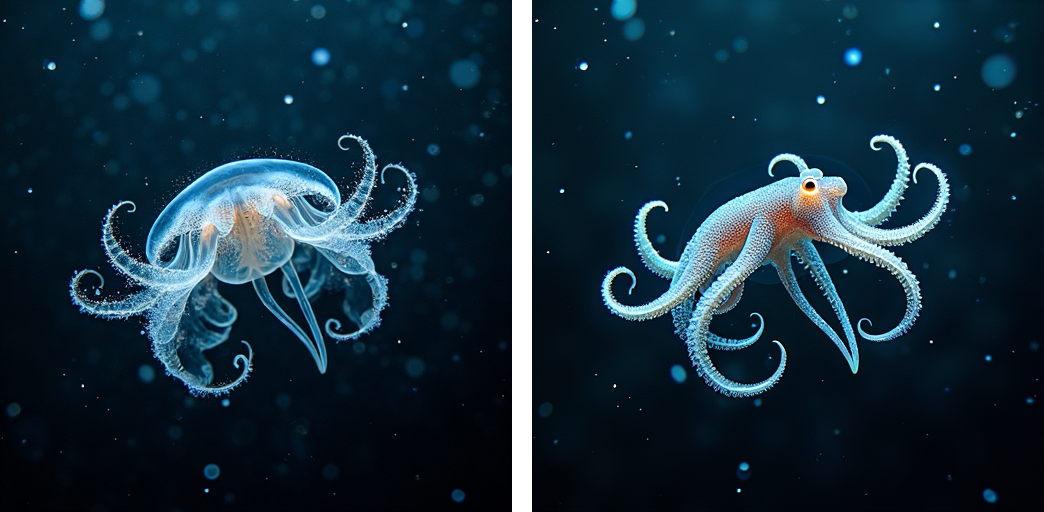}
    \caption{Comb jelly undulating in bioluminescent threads vs 'larval octopus undulating in bioluminescent threads, pigeons startled}
  \end{subfigure}%
  \hfill
  \begin{subfigure}[t]{\subfigwidth}
    \includegraphics[width=\textwidth]{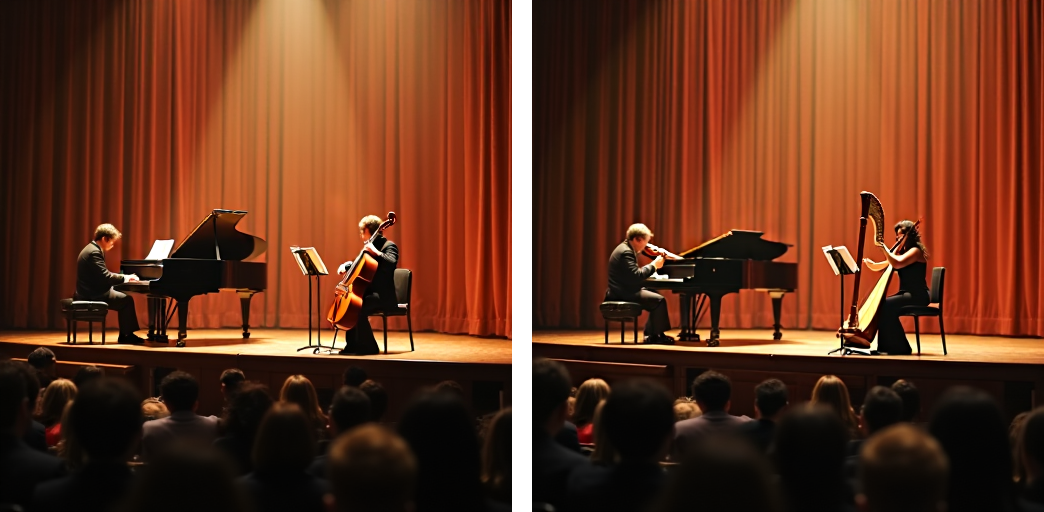}
    \caption{Concert stage with a pianist and a cellist performing in counterpoint vs Concert stage with a harpist and a flutist performing in counterpoint}
  \end{subfigure}
  
  \vspace{5pt} 
  
  \begin{subfigure}[t]{\subfigwidth}
    \includegraphics[width=\textwidth]{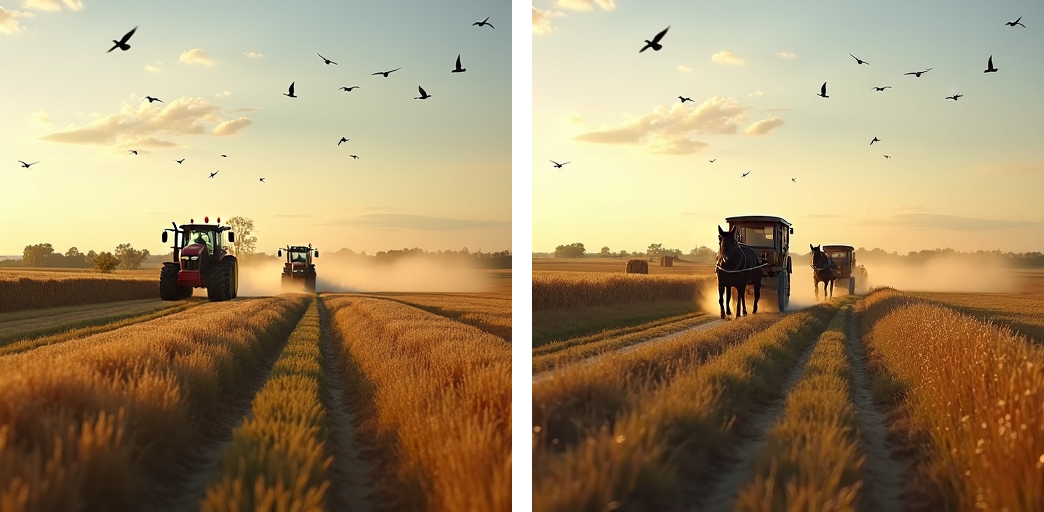}
    \caption{Farm lane with two tractors rolling past hay bales vs Farm lane with two horses pulling wagons past hay bales}
  \end{subfigure}%
  \hfill
  \begin{subfigure}[t]{\subfigwidth}
    \includegraphics[width=\textwidth]{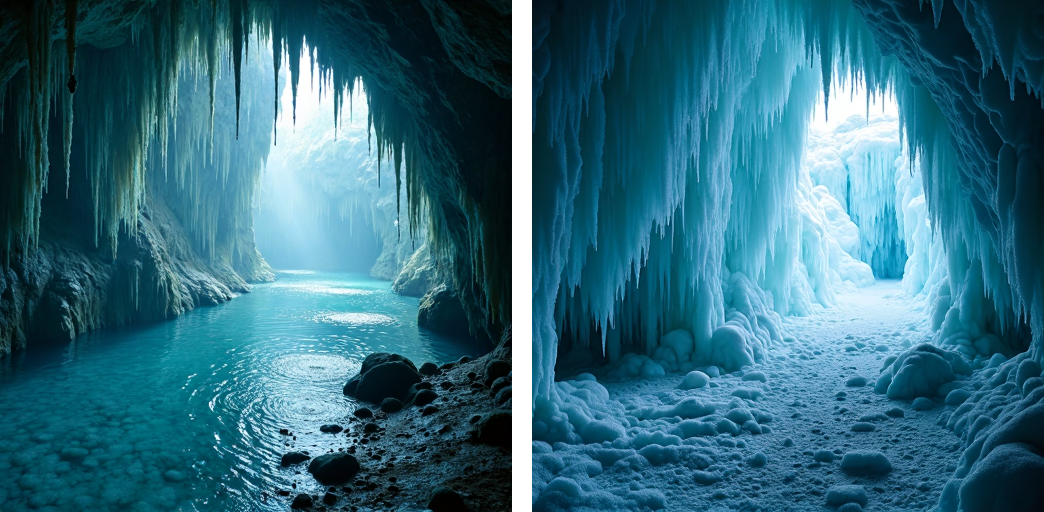}
    \caption{Limestone cave chamber vs Glacial ice cave chamber}
  \end{subfigure}%
  \hfill
  \begin{subfigure}[t]{\subfigwidth}
    \includegraphics[width=\textwidth]{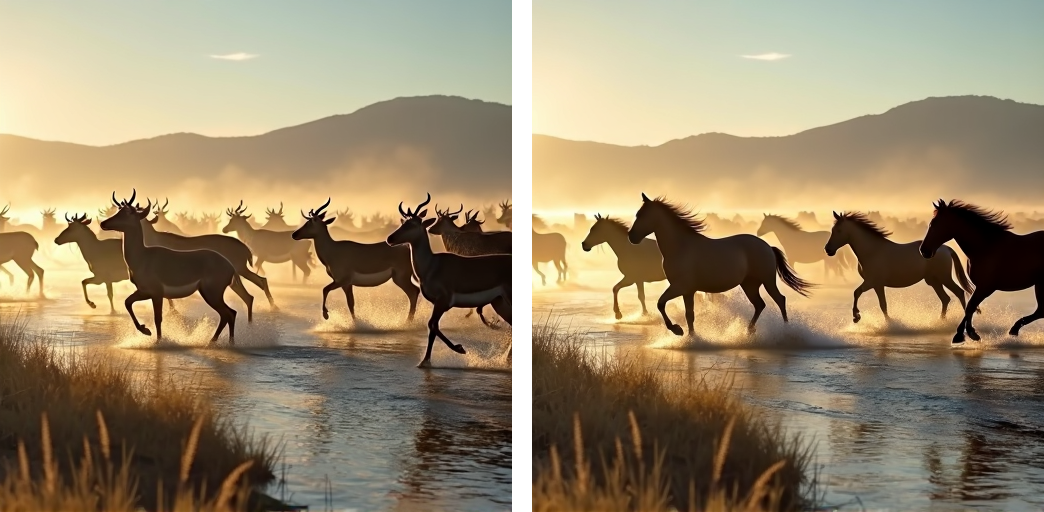}
    \caption{Open steppe with a sweeping herd of antelope crossing a river braid vs Open steppe with a sweeping herd of wild horses crossing a river braid}
  \end{subfigure}
  
  \vspace{5pt}
  
  \begin{subfigure}[t]{\subfigwidth}
    \includegraphics[width=\textwidth]{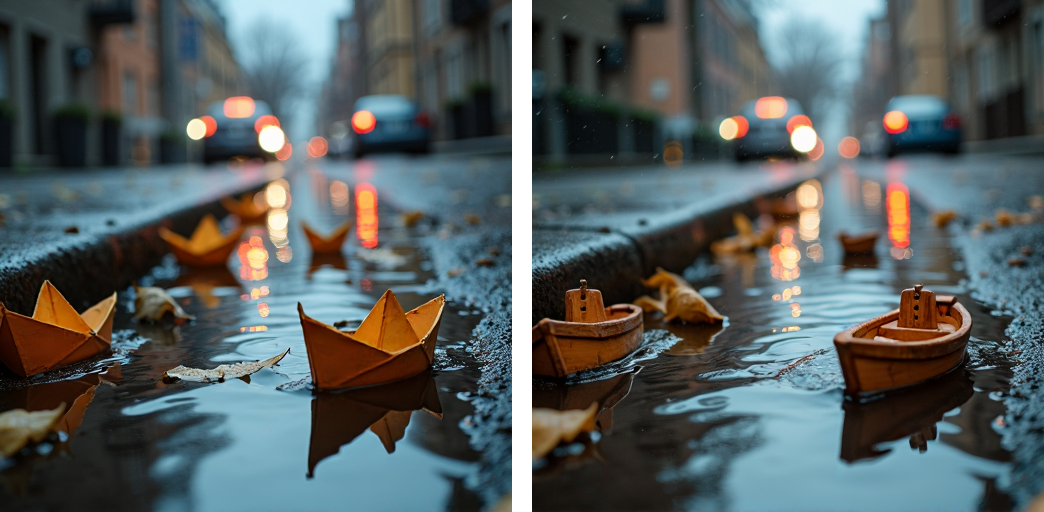}
    \caption{Rainy gutter stream with paper boats drifting past curb leaves vs Rainy gutter stream with wooden toy boats drifting past curb leaves}
  \end{subfigure}%
  \hfill
  \begin{subfigure}[t]{\subfigwidth}
    \includegraphics[width=\textwidth]{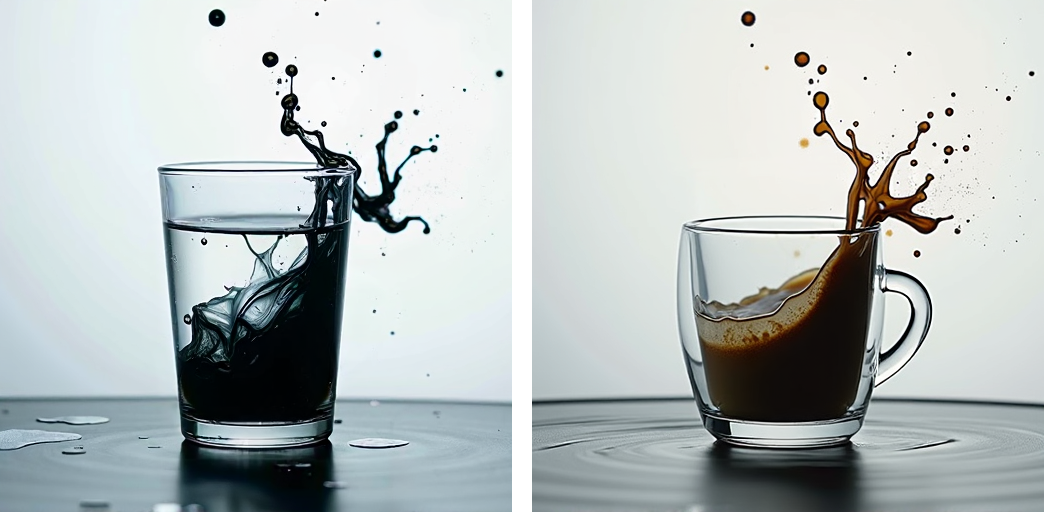}
    \caption{Art tabletop with ink plumes blooming in a paper marbling bath vs 'Art tabletop with paint plumes blooming in a paper marbling bath}
  \end{subfigure}%
  \hfill
  \begin{subfigure}[t]{\subfigwidth}
    \includegraphics[width=\textwidth]{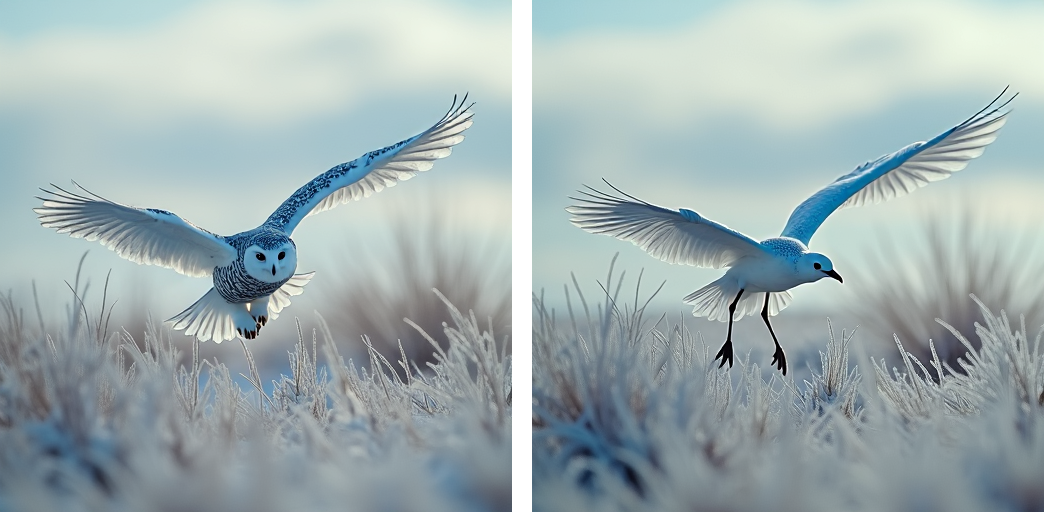}
    \caption{Tundra thermals with a snowy owl hovering above frost grass vs Tundra thermals with a frost wyvern hovering above frost grass}
  \end{subfigure}
  
  \caption{Aligned image generation results}
  \label{fig:grid}
\end{figure*}

Additional examples of applying our method to aligned image generation are presented in Figure \ref{fig:grid}. 

\section{Additional 3D generation results}
\begin{figure*}
     \centering
     \newcolumntype{C}{>{\centering\arraybackslash}m{0.2\linewidth}}
     \begin{tabular}{>{\centering\arraybackslash}m{3.0em} C C C C} 
          animal: &
         \includegraphics[width=\linewidth]{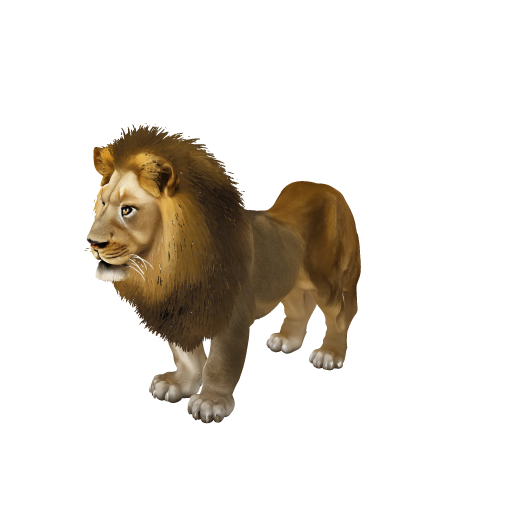} &
         \includegraphics[width=\linewidth]{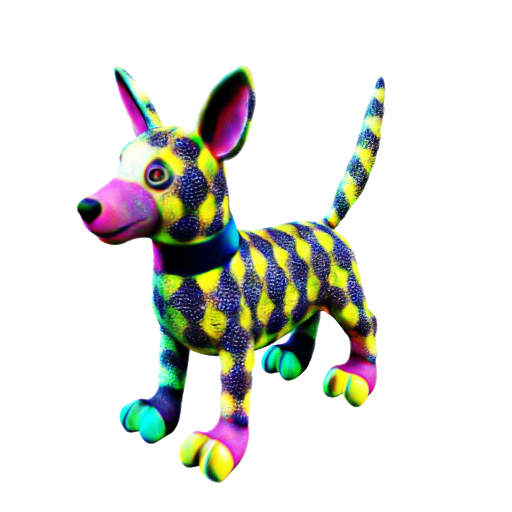} &
         \includegraphics[width=\linewidth]{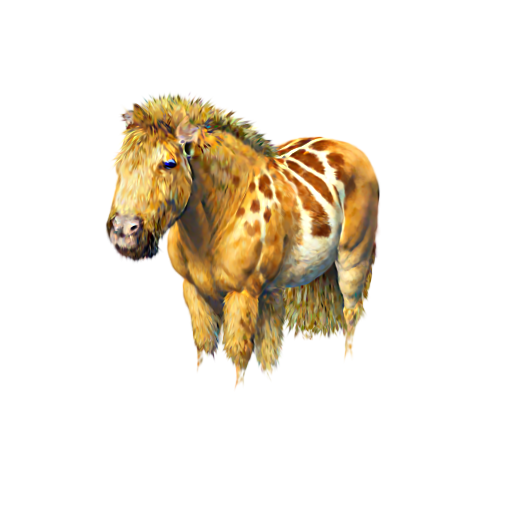} &
         \includegraphics[width=\linewidth]{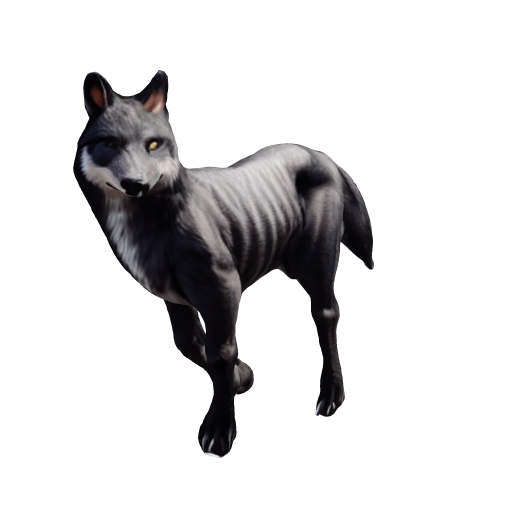} \\[2ex]
         lego animal: &
         \includegraphics[width=\linewidth]{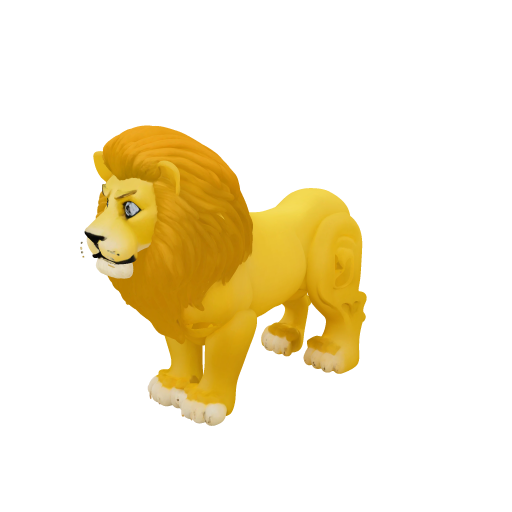} &
         \includegraphics[width=\linewidth]{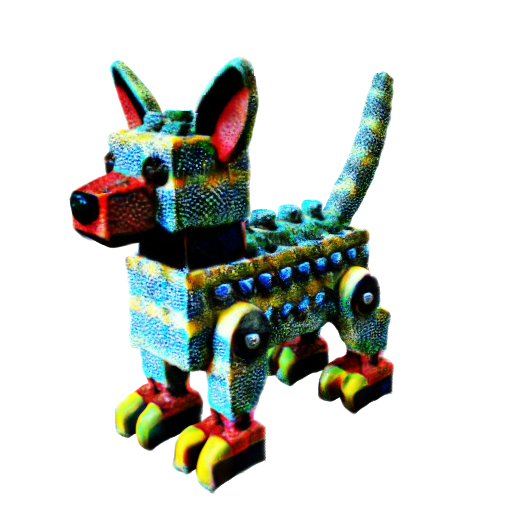} &
         \includegraphics[width=\linewidth]{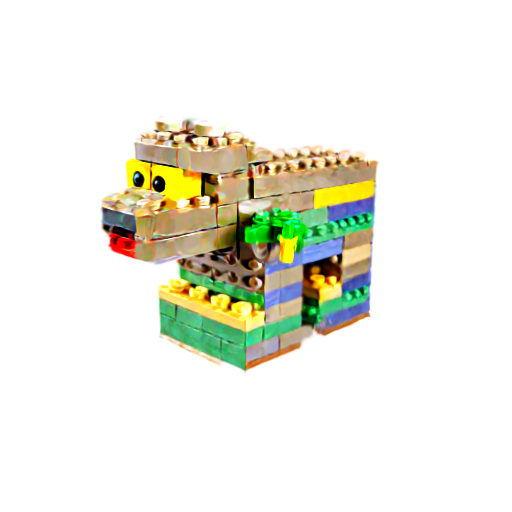} &
         \includegraphics[width=\linewidth]{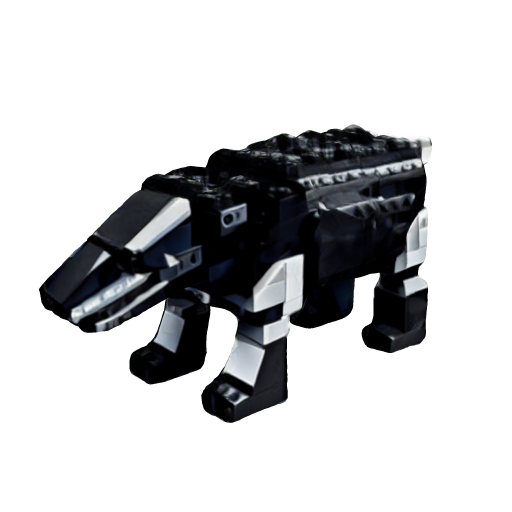} \\[2ex]
         atakebune ship: &
         \includegraphics[width=\linewidth]{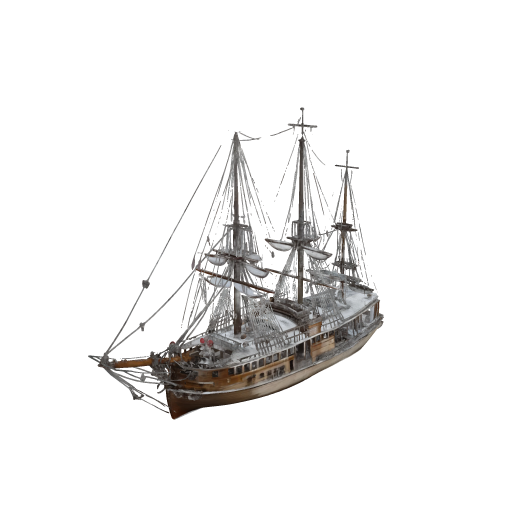} &
         \includegraphics[width=\linewidth]{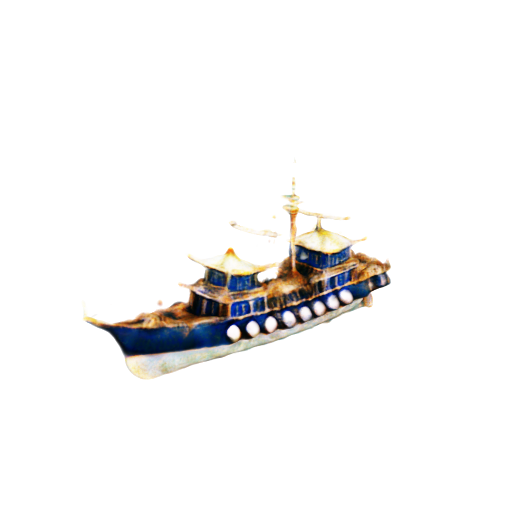} &
         \includegraphics[width=\linewidth]{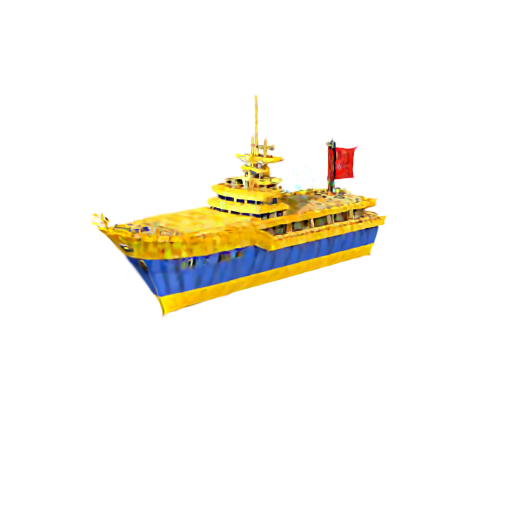} &
         \includegraphics[width=\linewidth]{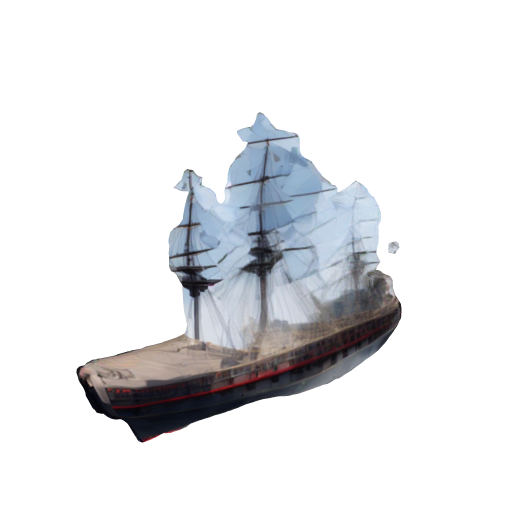} \\[2ex]
         modern yacht: &
         \includegraphics[width=\linewidth]{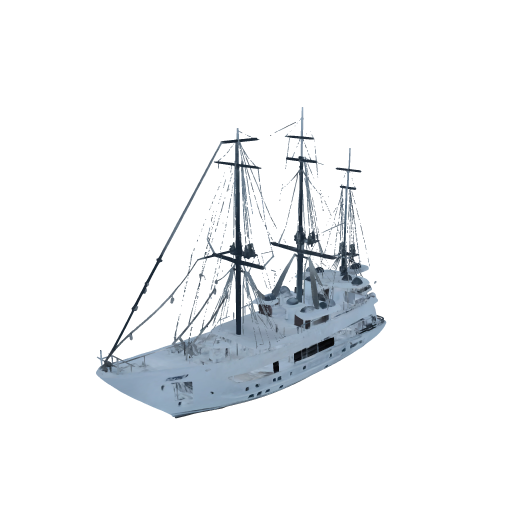} &
         \includegraphics[width=\linewidth]{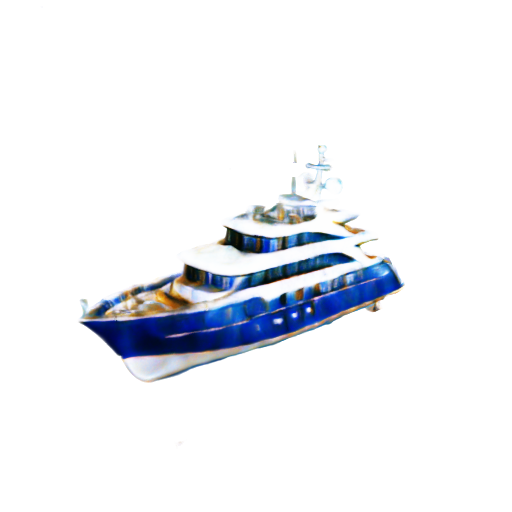} &
         \includegraphics[width=\linewidth]{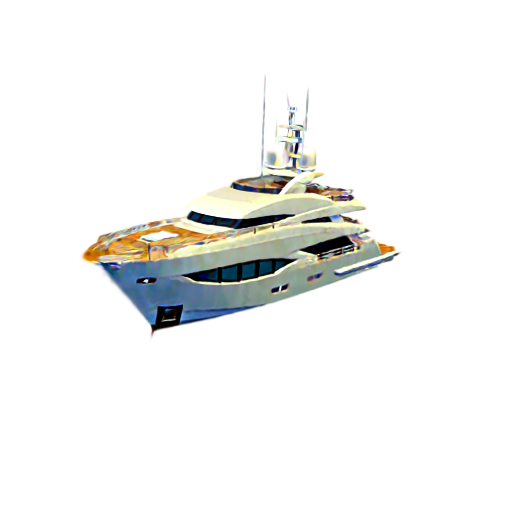} &
         \includegraphics[width=\linewidth]{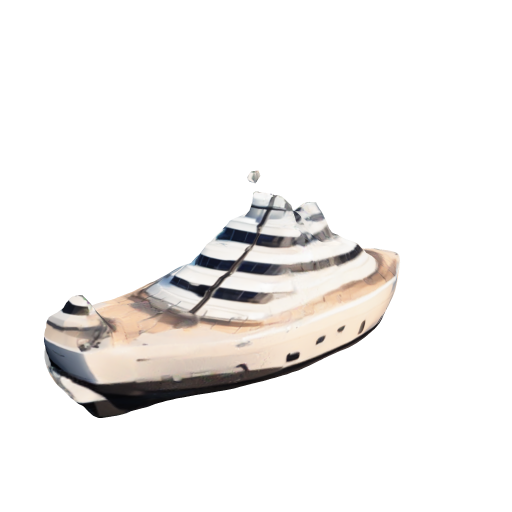} \\[2ex]
         bird animal: &
         \includegraphics[width=\linewidth]{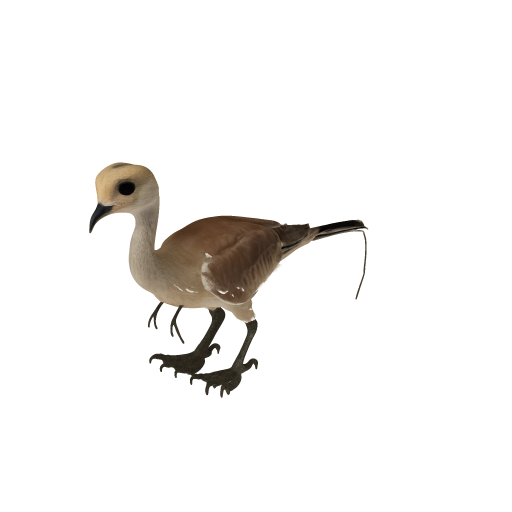} &
         \includegraphics[width=\linewidth]{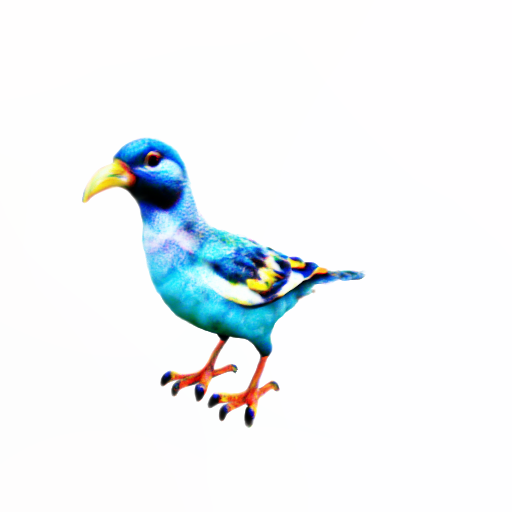} &
         \includegraphics[width=\linewidth]{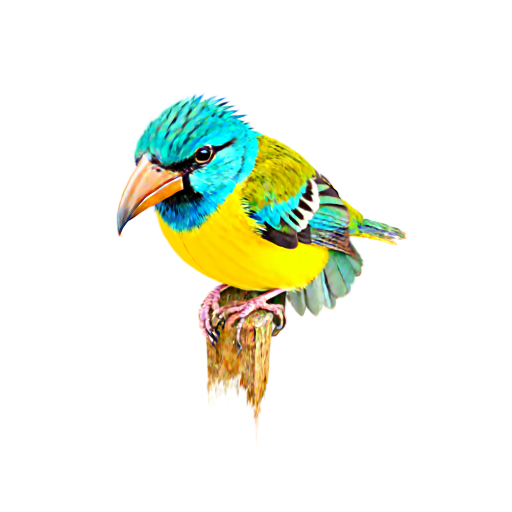} &
         \includegraphics[width=\linewidth]{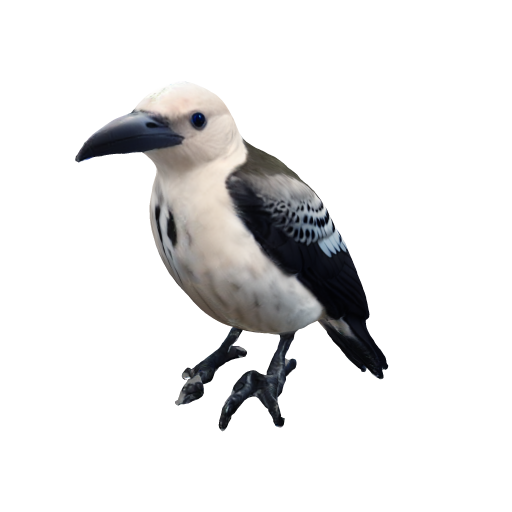} \\[2ex]
         dinosaur animal: &
         \includegraphics[width=\linewidth]{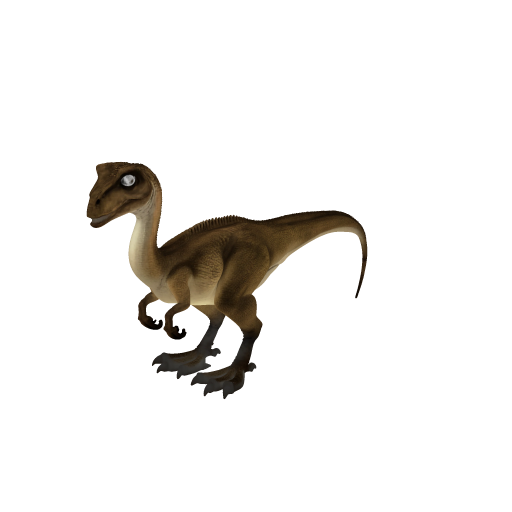} &
         \includegraphics[width=\linewidth]{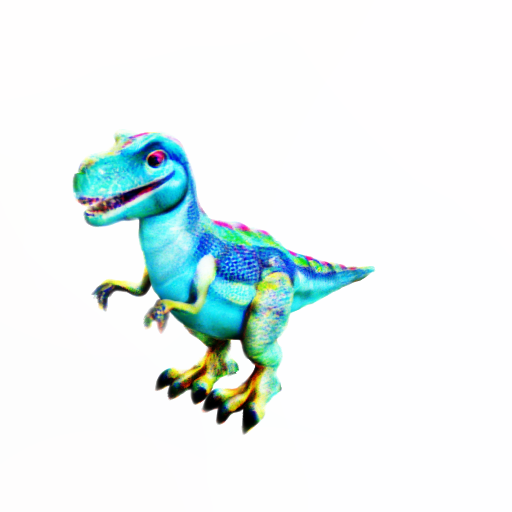} &
         \includegraphics[width=\linewidth]{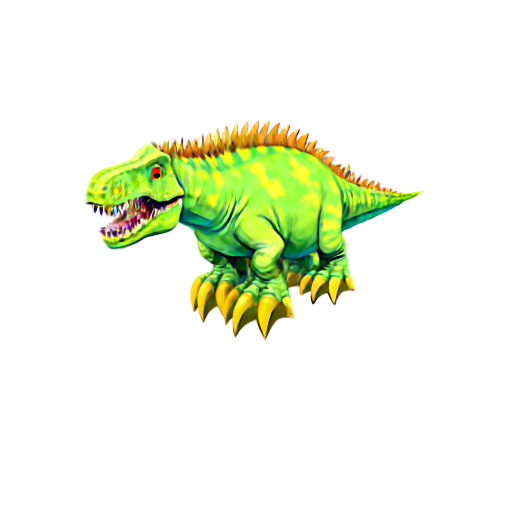} &
         \includegraphics[width=\linewidth]{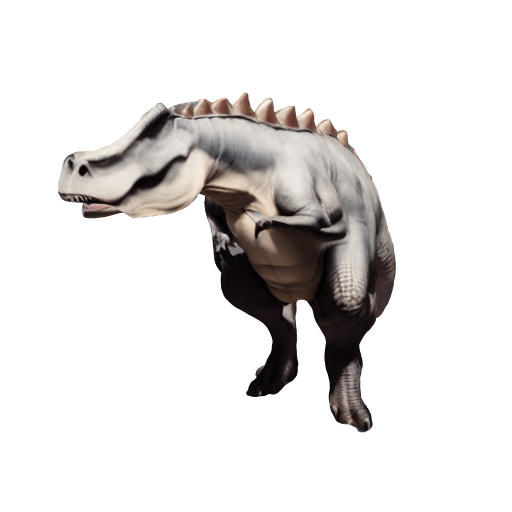} \\[2ex]
         & \textbf{Ours} & \textbf{A3D} & \textbf{LucidDreamer} & \textbf{MVEdit} \\
     \end{tabular}
     \caption{Visualization of 3D results in four different methods}
     \label{fig:3d_results}
 \end{figure*}

Figure \ref{fig:3d_results} provides additional examples of 3D generation results obtained with our method, along with a quantitative comparison against competing methods.


\section{Metrics details}

We primarily address the question of whether alignment can be achieved between the source image and the generated target image when the source prompt is modified into the target prompt. This process is evaluated based on two main criteria: 1) whether the generated target image accurately corresponds to the target prompt, and 2) whether proper alignment is maintained between the source image and the generated target image.

\subsection{2D Metrics} For 2D images, we employ the DIFT metric to evaluate the similarity between two images.
\[
S_{DIFT} = \frac{1}{2N} \sum_{i=1}^{N} \frac{\| F_A(P_i^A) - P_i^A \|_2}{\sigma_{P_A}} + \frac{\| F_B(P_i^B) - P_i^B \|_2}{\sigma_{P_B}},
\]
By uniformly sampling the masked image, we extract a dense point cloud of the object, which is then used to locate the most corresponding target point cloud in the aligned image.
The alignment between the two images is assessed by computing the L2 distance between corresponding points in the two point clouds.
To account for the influence of the object scale on distance calculation, we normalize the point clouds based on the size of the object. Additionally, to mitigate the impact of outlier points on the overall point cloud distance calculation, a distance threshold of 50 pixels is applied. To evaluate text-image alignment, we initially consider using the CLIP score as a metric. 
However, due to its limited interpretability, we adopt the evaluation approach from T2I-CompBench++, which employs the GPT-4o model to assess the alignment between text and images. 
This method, based on a multimodal large language model, not only provides a score but also offers textual explanations justifying the rating. 
To better highlight performance differences across methods, we uniformly scale up the original scores. 
We provide full results for different scenes in Table~\ref{sup:tab:full_2D_metrics}, our method demonstrates superior image alignment compared to Qwen and RF-Inversion. 
Although it slightly underperforms in text-image consistency, the difference is visually negligible.
\begin{table*}[t]
    \centering
    \caption{2D Metrics.}
    \includegraphics[scale=0.33]{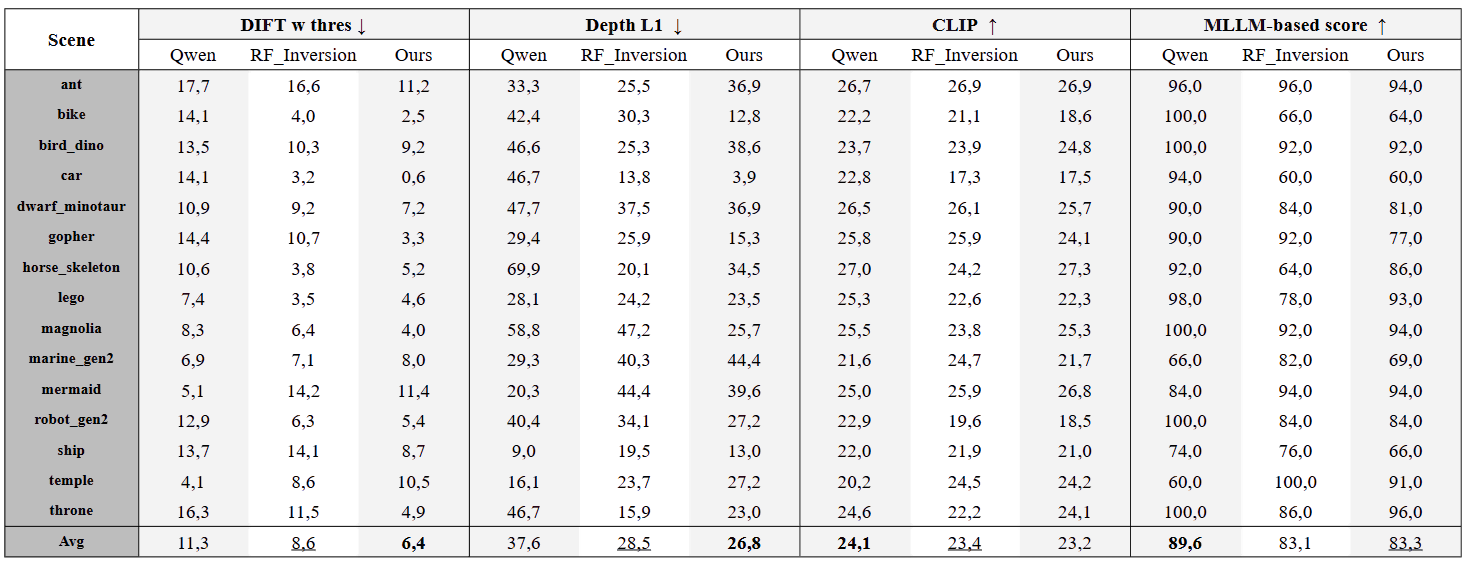}
    \label{sup:tab:full_2D_metrics}
\end{table*}   
\subsection{3D Metrics} 
Analogous to the evaluation of 2D images, we also assess 3D objects using the DIFT metric. 
The key distinction lies in our preliminary step of rendering each 3D object from 120 different viewpoints, resulting in 120 corresponding 2D images. 
The DIFT metric is then computed using the images rendered from both the source and target 3D objects. 
To evaluate text-image alignment, we compute the CLIP score using the rendered 2D images. 
As shown in Table~\ref{sup:tab:full_three_d_metrics}, with full results split into different scenes our method achieves superior image alignment compared to all competing methods.
In addition, the GPTEval3D metric was employed to assess performance across six dimensions: text-asset alignment, 3D plausibility, text-geometry alignment, texture Details, geometry details, and overall.

\subsection{Video Metrics}
The goal of video generation metrics is to evaluate all aspects of video pair generation: not only how well the geometry of the scene is preserved, but also how well the resulting videos look, their similarity to the prompt, etc.
We use 3 metrics in total, which cover all the requirements we want from video generation:
\begin{enumerate}
    \item \textbf{VLM evaluation}. We ask multi-modal LLM(gpt-4o) to give scores to 3 aspects: prompt following, edit quality, and background consistency. 
    Every aspect might give 3 points in total. \cite{ju2025editverseunifyingimagevideo}
    \item \textbf{DiNO consistency}. 
    We evaluate the temporal consistency of the whole video by calculating the cosine similarity between DINO features of the consecutive frames. 
    Instead of considering all DINO features, we compare only 'cls' tokens, which accumulate the overall semantics of the frames. \cite{ju2025editverseunifyingimagevideo} 
    \item \textbf{Depth difference}. 
    When generating pairs of videos, we want the geometry of the scene within pairs to be as close as possible as the reflection of the scene structure.
    We use the DepthAnythingv2 \cite{depth_anything_v2} model to estimate per-pixel depth and then calculate the difference between the depths of the corresponding pixels on the source and target videos.
\end{enumerate}
\begin{table*}[t]
    \centering
    \caption{3D Metrics.}
    \includegraphics[scale=0.4]{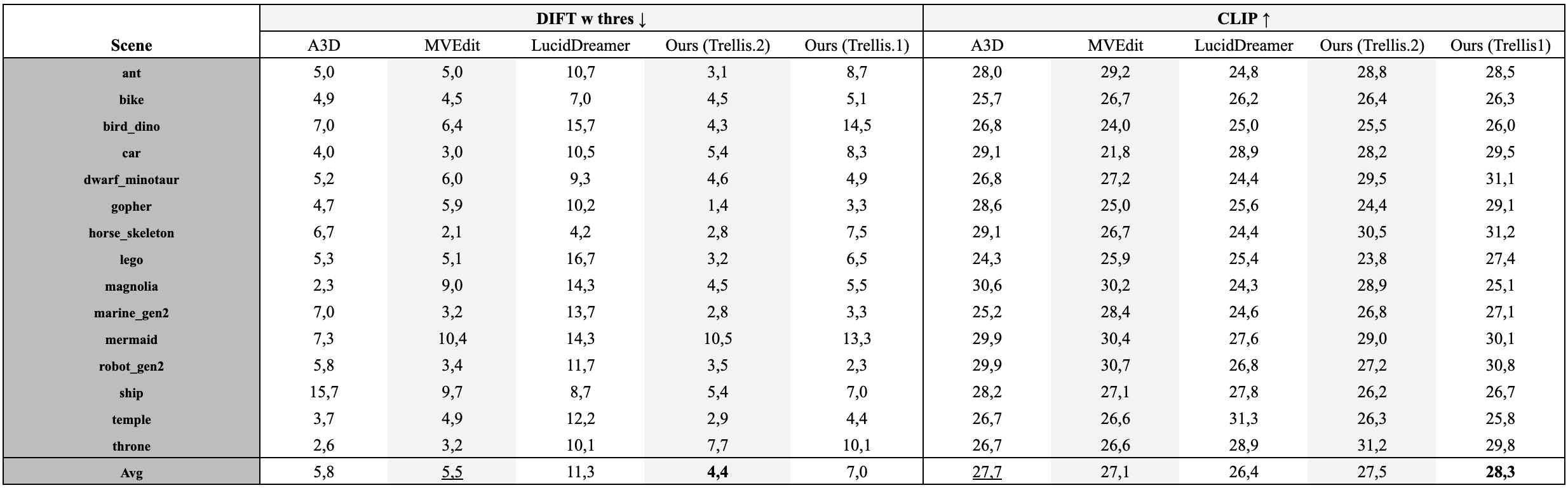}
    \label{sup:tab:full_three_d_metrics}
\end{table*}    
We present the full results for all the scenes in the Table~\ref{sup:tab:video_metrics}.

\section{Experiment details}

Our method depends on several schedules described in the Method section, such as the schedule of consistency scales $w_t$ in smoothness regularization and the schedule of $p(\alpha_i)$. 
We found that sampling only four $\alpha$ points is sufficient for approximation of the integral in Equation~\ref{sup:eq:integral_solution} to obtain plausible transitions between two objects in all three modalities.
We also found that the best scheduling for $w_t$ is a piecewise constant non-increasing function with several discontinuities that reduces the effect of regularization as the denoising timestep increases. This holds for all modalities. 

\subsection{Generating aligned images}

We use the following  $w_t$ schedule and $p(\alpha)$ density in our image generation experiments.

$$
w_t = 
\begin{cases}
0.7 & \text{if } t < 7 \\
0.5 & \text{if } t = 7 \\
0.4 & \text{if } t = 8 \\
0.1 & \text{if } t > 8
\end{cases}
$$
$$
p(\alpha) \propto 
\begin{cases}
p_{\mathcal{U}[0, 0.1)}(\alpha) & \text{if } \alpha \in [0,0.1) \\
0 & \text{if } \alpha \in [0.1, 0.3) \\
0.87 \: p_{\mathcal{U}[0.3, 0.5)}(\alpha) & \text{if } \alpha \in [0.3, 0.5) \\
0.5 & \text{if } \alpha = 0.5 \\
0.87 \: p_{\mathcal{U}(0.5, 0.7)}(\alpha) & \text{if } \alpha \in (0.5, 0.7) \\
0 & \text{if } \alpha \in [0.7, 0.9) \\
p_{\mathcal{U}[0.9, 1]}(\alpha) & \text{if } \alpha \in [0.9, 1.0], \\
\end{cases}
$$

\subsection{Generating aligned videos}
We use the following  $w_t$ schedule and $p(\alpha)$ density in our video generation experiments.

$$
w_t = 
\begin{cases}
0.5 & \text{if } t < 7 \\
0.4 & \text{if } t = 7 \\
0.3 & \text{if } t = 8 \\
0.1 & \text{if } t > 8
\end{cases}
$$
$$
p(\alpha) \propto 
\begin{cases}
p_{\mathcal{U}[0, 0.1)}(\alpha) & \text{if } \alpha \in [0,0.1) \\
0 & \text{if } \alpha \in [0.1, 0.3) \\
0.25 \: p_{\mathcal{U}[0.3, 0.5)}(\alpha) & \text{if } \alpha \in [0.3, 0.5) \\
0.15 & \text{if } \alpha = 0.5 \\
0.25 \: p_{\mathcal{U}(0.5, 0.7)}(\alpha) & \text{if } \alpha \in (0.5, 0.7) \\
0 & \text{if } \alpha \in [0.7, 0.9) \\
p_{\mathcal{U}[0.9, 1]}(\alpha) & \text{if } \alpha \in [0.9, 1.0], \\
\end{cases}
$$

\subsection{Generating aligned 3D objects}

We use the Trellis model as the backbone text-to-3D Flow Matching model.
This model consists of two main parts.
The first Flow Matching model is used to densely denoise and obtain a structured latent - sparse geometry representation. 
When the geometry is fixed, the second model is used to denoise the structured latents to obtain the details for the earlier estimated geometry. 
As we are interested in aligning geometry only, we incorporated our method only in the dense denoising part. 

We use the following  $w_t$ schedule and $p(\alpha)$ density in our 3D generation experiments.
In this way, the approximation of the \ref{sup:eq:integral_solution} can be achieved by sampling single points from the probability regions with non-zero density $p(\alpha)$ with the corresponding summarized probabilities.

$$
w_t = 
\begin{cases}
0.7 & \text{if } t < 12 \\
0.05 & \text{if } t > 12
\end{cases}
$$
$$
p(\alpha) \propto 
\begin{cases}
p_{\mathcal{U}[0, 0.1)}(\alpha) & \text{if } \alpha \in [0,0.1) \\
0 & \text{if } \alpha \in [0.1, 0.3) \\
0.35 \: p_{\mathcal{U}[0.3, 0.5)}(\alpha) & \text{if } \alpha \in [0.3, 0.5) \\
0.5 & \text{if } \alpha = 0.5 \\
0.35 \: p_{\mathcal{U}(0.5, 0.7)}(\alpha) & \text{if } \alpha \in (0.5, 0.7) \\
0 & \text{if } \alpha \in [0.7, 0.9) \\
p_{\mathcal{U}[0.9, 1]}(\alpha) & \text{if } \alpha \in [0.9, 1.0], \\
\end{cases}
$$
\begin{table*}[t]
    \centering
    \caption{Video Metrics.}
    \includegraphics[scale=0.33]{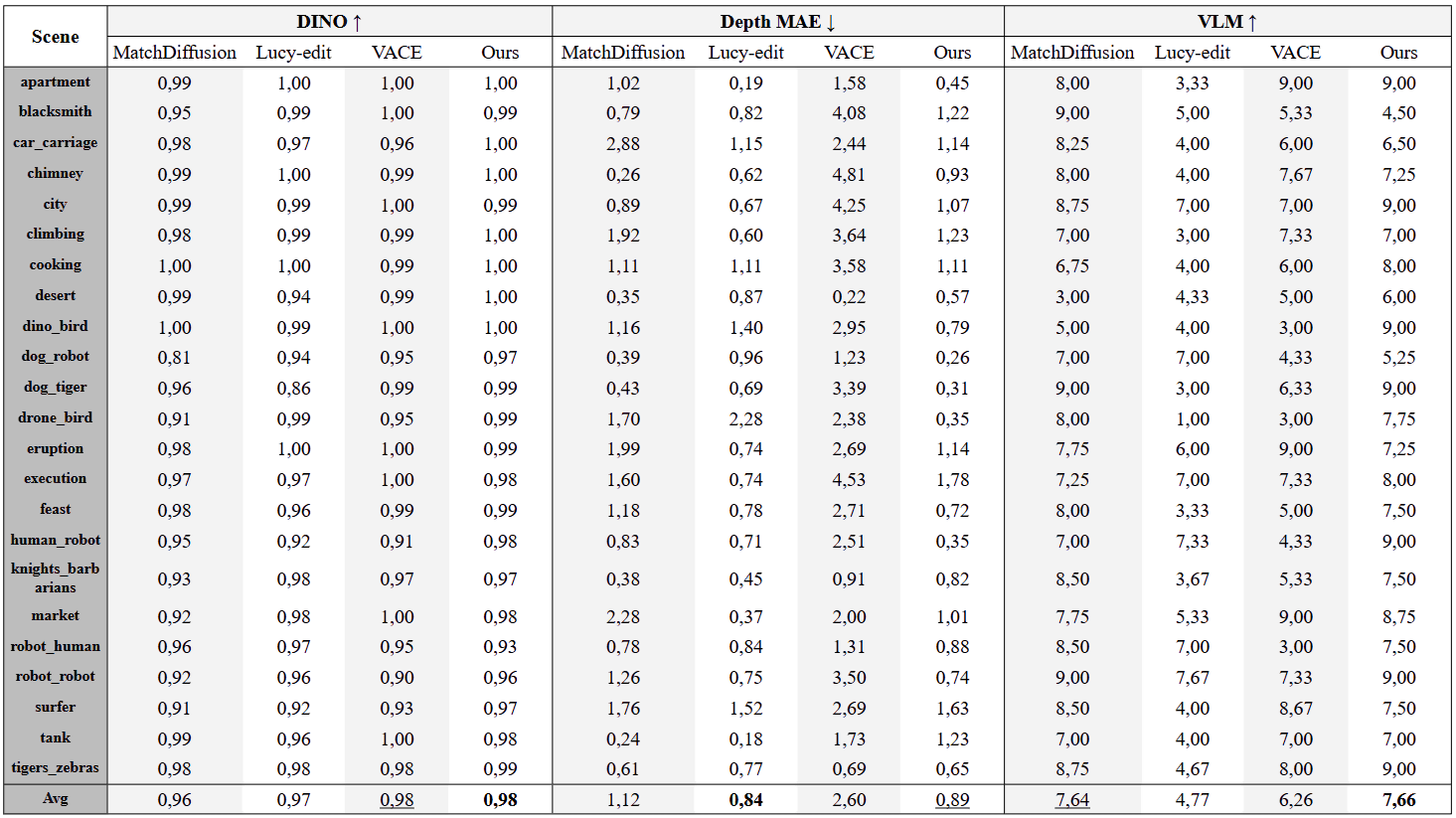}
    \label{sup:tab:video_metrics}
\end{table*}

\subsection{Generating aligned 3D objects with Trellis.2}
A more recent and more capable 3D generative model with structured latents than Trellis is Trellis.2. However, it supports only image-to-3D generation, so we introduced minor modifications to our method. All implementations described above were designed for text-conditioned generation and therefore relied on text-prompt interpolation. In contrast, image-conditioned generation requires aligned image prompts and interpolation in the image-token (or image-embedding) space.

To obtain aligned 3D objects with Trellis.2, we proceed in two stages. First, we use our Flux modification to generate a pair of aligned images. Second, we apply our Trellis.2 modification to reconstruct aligned 3D assets from these images. This pipeline closely follows the denoising loop described in the main paper; the only change is the conditioning interpolation step: instead of interpolating text embeddings, we interpolate image tokens.

Because Trellis.2 follows image prompts more strongly than Trellis.1 follows text prompts, we found it necessary to increase the anchoring regularization strength. The final hyperparameters are reported below:
$$
w_t = 
\begin{cases}
0.9 & \text{if } t < 12 \\
0.8 & \text{if } t > 12
\end{cases}
$$
$$
p(\alpha) \propto 
\begin{cases}
p_{\mathcal{U}[0, 0.1)}(\alpha) & \text{if } \alpha \in [0,0.1) \\
0 & \text{if } \alpha \in [0.1, 0.3) \\
2.5 \: p_{\mathcal{U}[0.3, 0.5)}(\alpha) & \text{if } \alpha \in [0.3, 0.5) \\
1.5 & \text{if } \alpha = 0.5 \\
2.5 \: p_{\mathcal{U}(0.5, 0.7)}(\alpha) & \text{if } \alpha \in (0.5, 0.7) \\
0 & \text{if } \alpha \in [0.7, 0.9) \\
p_{\mathcal{U}[0.9, 1]}(\alpha) & \text{if } \alpha \in [0.9, 1.0], \\
\end{cases}
$$
\subsection{Prompts}
For video and 3D experiments, we have to use the detailed versions of the prompts, since modern models tend to perform much better with long and detailed prompts than with short ones.
We provide the full prompts for 3D experiments in Table~\ref{sup:tab:3d_prompts} and full prompts for video experiments in Table~\ref{sup:tab:video_promtps}. 

\begin{table*}[t]
    \centering
    \caption{Video Prompts.}
    \includegraphics[scale=0.33]{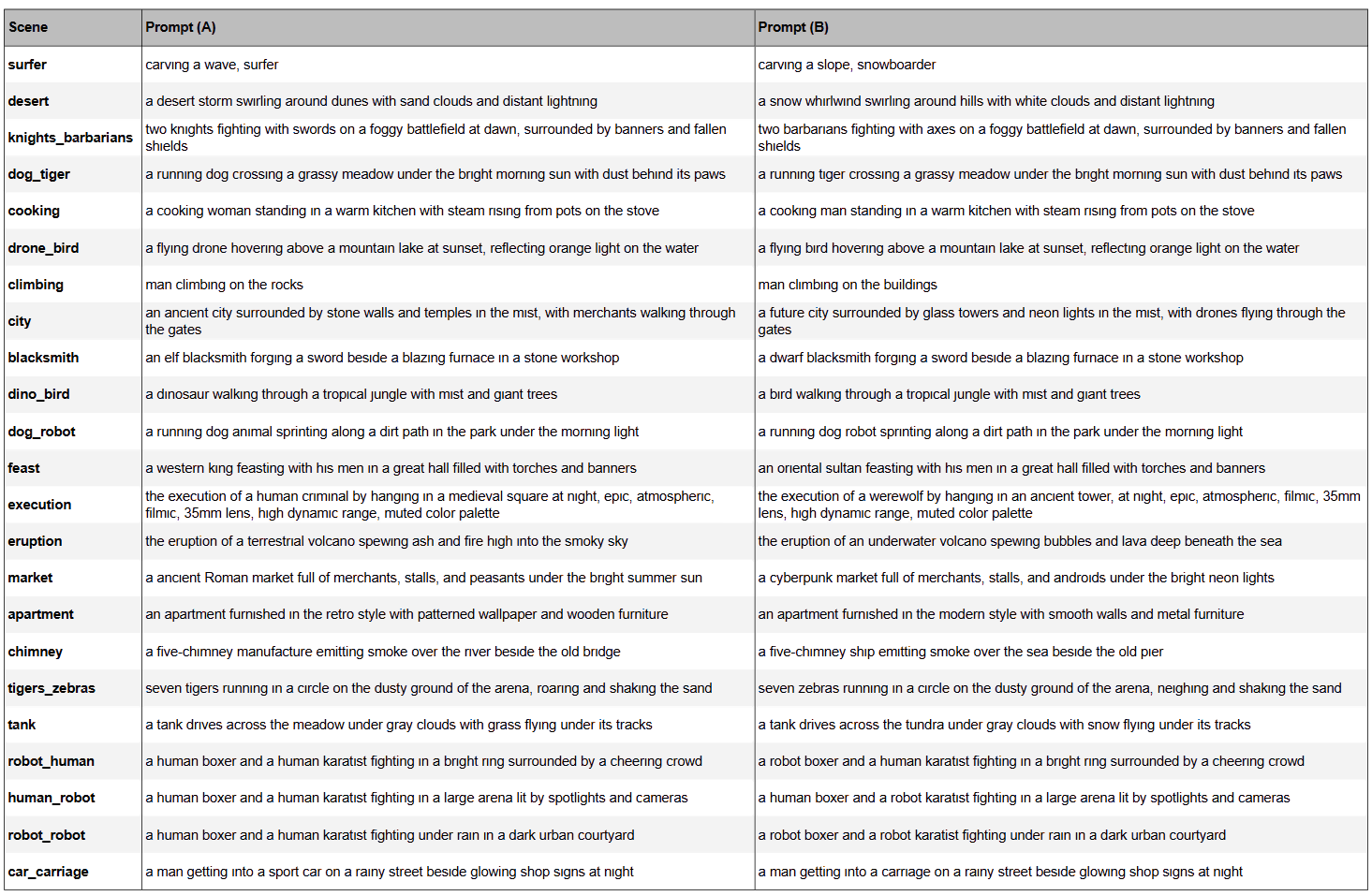}
    \label{sup:tab:video_promtps}
\end{table*}

\begin{table*}[t]
    \centering
    \caption{3D Prompts.}
    \includegraphics[scale=0.33]{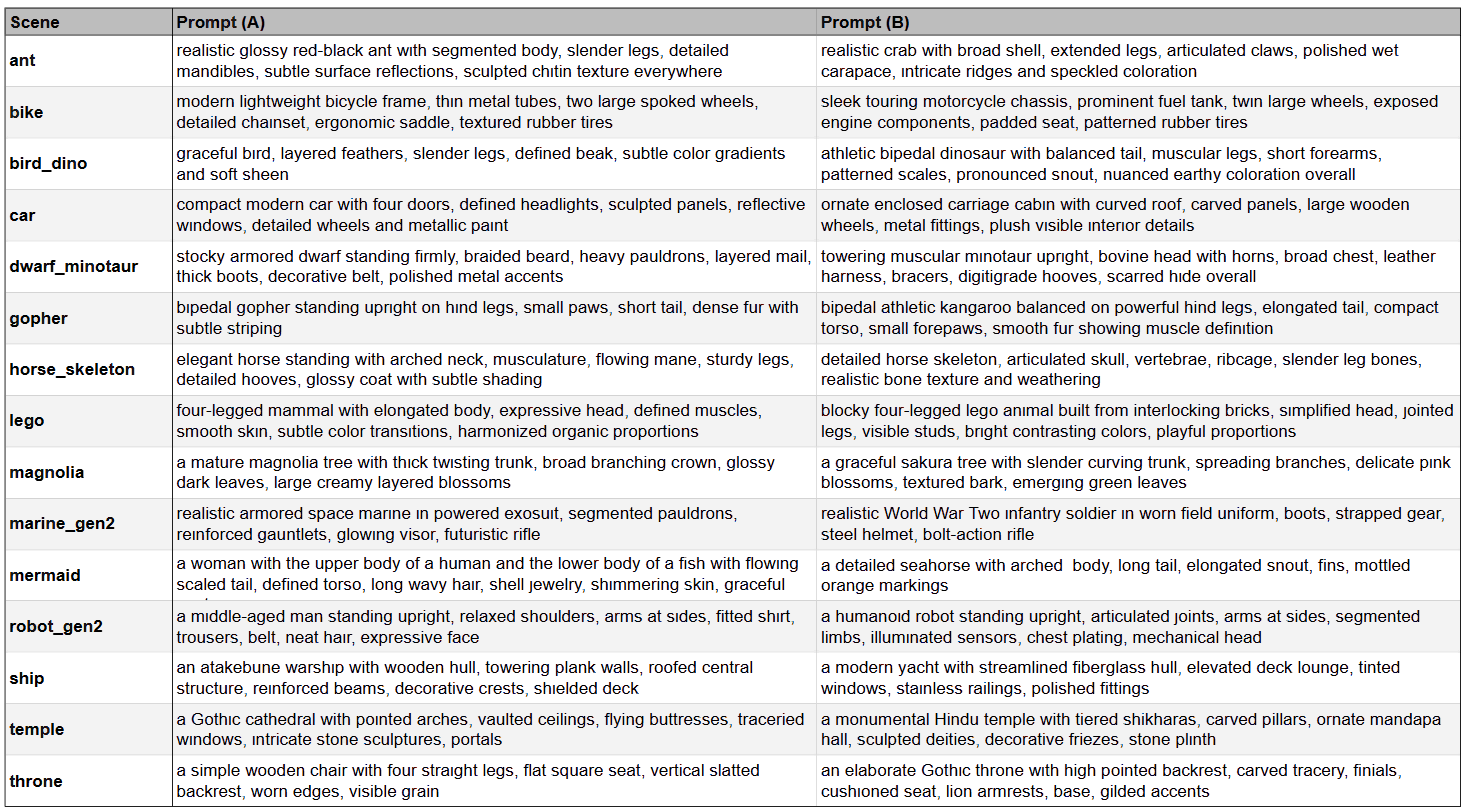}
    \label{sup:tab:3d_prompts}
\end{table*}

\end{document}